\documentclass[letterpaper, 10 pt, journal, twoside]{IEEEtran}
\newcommand{\etal}{{\it{et al}.} }

\usepackage{subfigure}
\usepackage{amsmath}
\usepackage{amssymb}
\usepackage{xcolor}
\usepackage{multirow}
\usepackage{tabu}
\usepackage{pbox}
\usepackage{graphicx}
{\begin{list}               %    with flush left bullets.
    {$\bullet$ \hfill}{
        \setlength{\leftmargin}{\parindent}
        \setlength{\parsep}{0.04\baselineskip}
        \setlength{\itemsep}{0.5\parsep}
        \setlength{\labelwidth}{\leftmargin}
        \setlength{\labelsep}{0em}}
    }
{\end{list}}

\providecommand{\cref}[1]{Chapter~\ref{#1}}
\providecommand{\sref}[1]{Section~\ref{#1}}
\providecommand{\fref}[1]{Fig.~\ref{#1}}
\providecommand{\tref}[1]{Table~\ref{#1}}

\providecommand{\norm}[1]{\lVert#1\rVert}

\renewcommand{\vec}[1]{\ensuremath{\boldsymbol{#1}}}
\providecommand{\mat}[1]{\ensuremath{\boldsymbol{#1}}}

\providecommand{\calL}{\mathcal{L}}

\providecommand{\calR}{\mathcal{R}}

\providecommand{\mI}{\mat{I}}

\providecommand{\vc}{\vec{c}}
\providecommand{\ve}{\vec{e}}

\title{Image-to-Image Translation-based Data Augmentation for Robust EV Charging Inlet Detection
}

\author{Yeonjun Bang$^{1}$, Yeejin Lee$^{2}$, and Byeongkeun Kang$^{3}$  % 
\thanks{$^{1}$Yeonjun Bang and $^{3}$Byeongkeun Kang are with the Department of Electronic Engineering, Seoul National University of Science and Technology, South Korea
        {\tt\footnotesize yjbang0529@seoultech.ac.kr, byeongkeun.kang@seoultech.ac.kr}}%
\thanks{$^{2}$Yeejin Lee is with the Department of Electrical and Information Engineering, Seoul National University of Science and Technology, South Korea
        {\tt\footnotesize yeejinlee@seoultech.ac.kr}}
}

\begin{document}
\maketitle
% \thispagestyle{empty}
% \pagestyle{empty}

%%%%%%%%%%%%%%%%%%%%%%%%%%%%%%%%%%%%%%%%%%%%%%%%%%%%%%%%%%%%%%%%%%%%%%%%%%%%%%%%
\begin{abstract}
This work addresses the task of electric vehicle (EV) charging inlet detection for autonomous EV charging robots. Recently, automated EV charging systems have received huge attention to improve users' experience and to efficiently utilize charging infrastructures and parking lots. However, most related works have focused on system design, robot control, planning, and manipulation. Towards robust EV charging inlet detection, we propose a new dataset (EVCI dataset) and a novel data augmentation method that is based on image-to-image translation where typical image-to-image translation methods synthesize a new image in a different domain given an image. To the best of our knowledge, the EVCI dataset is the first EV charging inlet dataset. For the data augmentation method, we focus on being able to control synthesized images' captured environments (e.g., time, lighting) in an intuitive way. To achieve this, we first propose the environment guide vector that humans can intuitively interpret. We then propose a novel image-to-image translation network that translates a given image towards the environment described by the vector. Accordingly, it aims to synthesize a new image that has the same content as the given image while looking like captured in the provided environment by the environment guide vector. Lastly, we train a detection method using the augmented dataset. Through experiments on the EVCI dataset, we demonstrate that the proposed method outperforms the state-of-the-art methods. We also show that the proposed method is able to control synthesized images using an image and environment guide vectors. 
\end{abstract}

%\begin{IEEEkeywords}
%Deep learning for visual perception, computer vision for automation, data sets for robotic vision
%\end{IEEEkeywords}

%%%%%%%%%%%%%%%%%%%%%%%%%%%%%%%%%%%%%%%%%%%%%%%%%%%%%%%%%%%%%%%%%%%%%%%%%%%%%%%%

\section{INTRODUCTION}

\begin{figure}[!t] \begin{center}
\begin{minipage}{0.32\linewidth}
\centerline{\includegraphics[scale=0.155]{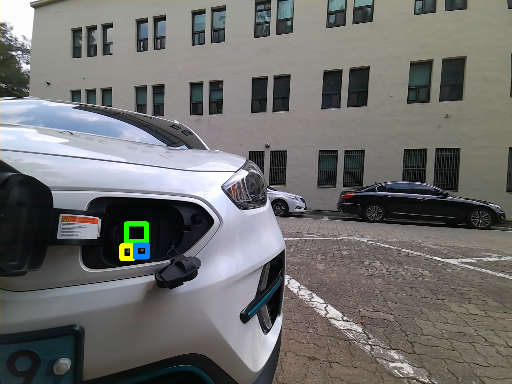}}
\end{minipage}
\begin{minipage}{0.32\linewidth}
\centerline{\includegraphics[scale=0.155]{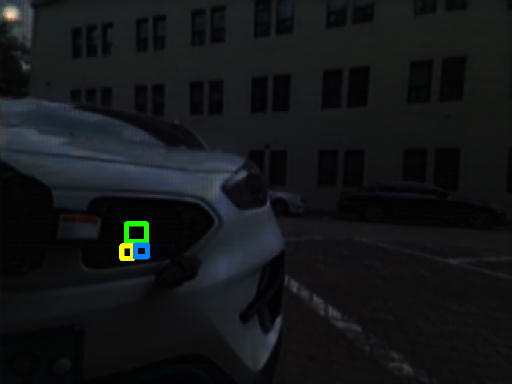}}
\end{minipage}
\begin{minipage}{0.32\linewidth}
\centerline{\includegraphics[scale=0.155]{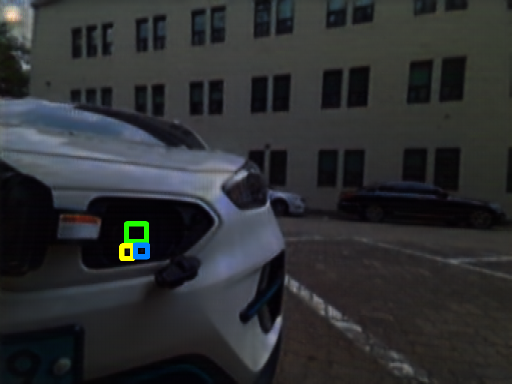}}
\end{minipage}
\\
\vspace{0.05cm}
\begin{minipage}{0.32\linewidth}
\centerline{\footnotesize{Original Image}}
\end{minipage}
\begin{minipage}{0.32\linewidth}
\centerline{\footnotesize{$\ve=(-0.8, -0.8, -0.8)$}}
\end{minipage}
\begin{minipage}{0.32\linewidth}
\centerline{\footnotesize{$\ve=(-0.5, -0.5, -0.5)$}}
\end{minipage}
\\
\vspace{0.05cm}
\begin{minipage}{0.32\linewidth}
\centerline{\includegraphics[scale=0.155]{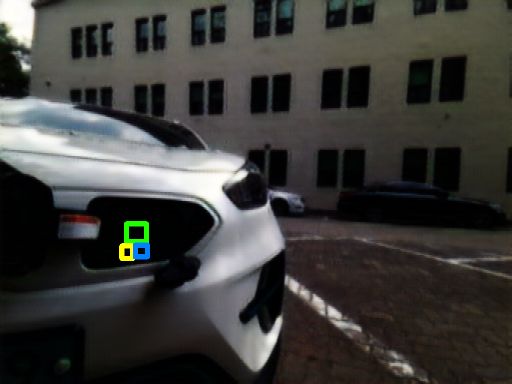}}
\end{minipage}
\begin{minipage}{0.32\linewidth}
\centerline{\includegraphics[scale=0.155]{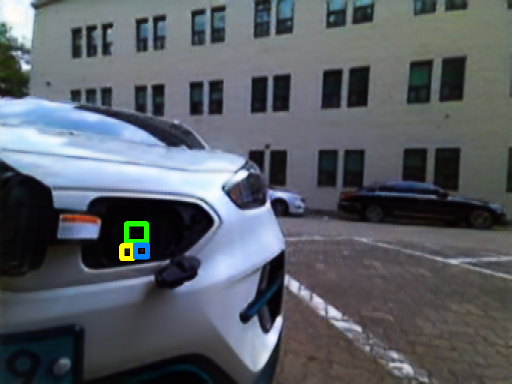}}
\end{minipage}
\begin{minipage}{0.32\linewidth}
\centerline{\includegraphics[scale=0.155]{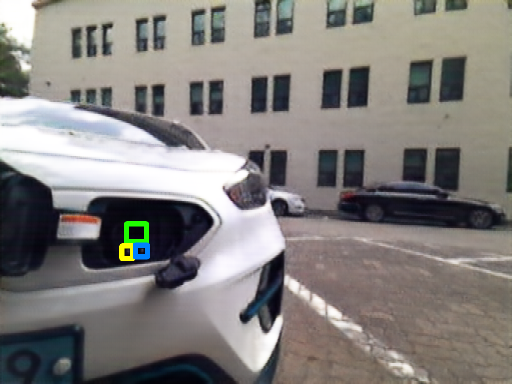}}
\end{minipage}
\\
\vspace{0.05cm}
\begin{minipage}{0.32\linewidth}
\centerline{\footnotesize{$\ve=(-0.5, 0.5, -0.5)$}}
\end{minipage}
\begin{minipage}{0.32\linewidth}
\centerline{\footnotesize{$\ve=(0., 0., 0.)$}}
\end{minipage}
\begin{minipage}{0.32\linewidth}
\centerline{\footnotesize{$\ve=(0.8, 0.8, -0.8)$}}
\end{minipage}
   \caption{Proposed image-to-image translation method synthesizes a new image by translating an original image towards the domain that is described by the proposed explainable environment vector. Top-left image shows an original image; Others show translated images using the provided environment vector $\ve$ shown at the bottom of each image.}
\label{fig:teaser}
\end{center}\end{figure}

Localizing the charging inlet of an electric vehicle (EV) is an essential task for autonomous EV charging robots to accurately plug-in a charging coupler. As the market share of EVs has recently increased explosively, charging infrastructures have also huge demands. Current charging stations are, however, inconvenient since human is required in-between charging a vehicle to another one. For instance, if a charger is coupled to a vehicle, to charge another one, a human needs to plug-out the charger from the former and to plug-in it into the latter. For fast EV chargers, this is even harder as power cables are heavier (around 10kg). Hence, to alleviate the inconvenience, autonomous EV charging robots have been proposed as an efficient and effective charging method~\cite{Behl2019, evChargingMiseikis2017, evChargingLong2019, evChargingLou2020, evChargingSun2018}. 

\begin{figure}[!t] \begin{center}
\begin{minipage}{0.49\linewidth}
\centerline{\includegraphics[scale=0.58]{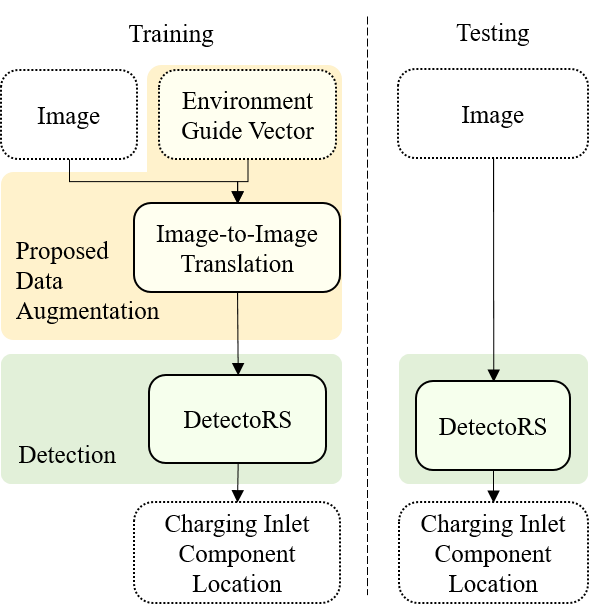}}
\end{minipage}
   \caption{Overview of the proposed framework. It shows the training and testing of the detection network utilizing the proposed data augmentation method.}
\label{fig:overview}
\end{center}\end{figure}

While a robot needs to locate the charging inlet of a vehicle to charge the EV automatically, very limited researches have focused on EV charging inlet localization. Accordingly, we were not able to find a public dataset. Moreover, to the best of our knowledge, most (if not all) experiments were conducted under lab environments without a real vehicle. Hence, we, first of all, propose the first public EV charging inlet localization dataset (EVCI dataset)\footnote{https://github.com/machinevision-seoultech/evci.}. Furthermore, to achieve robust localization, we propose a novel image-to-image translation network and employ the network for data augmentation. As image-to-image translation networks generate a synthesized image given an image, these networks can naturally increase the diversity of images in the dataset unless the synthesized image is exactly the same as the one in the dataset.
In the proposed translation network, to be able to guide synthesized images' captured environments, we propose to utilize explainable environment guide vectors. By this, we were able to generate various and explainable images given an image and varying environment vectors (see~\fref{fig:teaser}). Lastly, a state-of-the-art detection network is trained using the augmented dataset and is employed to achieve robust localization. The overall framework is shown in~\fref{fig:overview}.

The contributions of this paper are as follow: (1) We present, to the best of our knowledge, the first EV charging inlet localization dataset that is essential for the research related to autonomous EV charging robots; (2) To be able to control the conditions of synthesized images, we propose the explainable environment guide vector. By using the vector, we are able to synthesize various corresponding images given an image (see~\fref{fig:teaser}); (3) We propose a novel image-to-image translation network that utilizes the interpretable environment vector. The translation network is employed to augment training data; (4) We demonstrate the effectiveness of the proposed data augmentation method by training a state-of-the-art detection network using the augmented data.

\section{Related works}
\subsection{EV Charging Inlet Detection}
Miseikis \etal used a shape-based template matching algorithm to localize charging ports in a pair of stereo images under a lab environment~\cite{evChargingMiseikis2017}. Depth is then estimated using corresponding coordinates on the image pair. Given 3D coordinates, perspective transformation is estimated using a least-squares fit algorithm to obtain refined position and orientation. Behl \etal presented a proof-of-concept robot to automatically charge an EV~\cite{Behl2019}. They divided the robot system into three subsystems and evaluated each subsystem independently. Regarding charging inlet plug-in/out, they used a 3D printed charging inlet instead of a real vehicle. They localized the inlet on a captured image from a camera by using the Hough transform~\cite{houghTransform1972} and a template matching algorithm. Long \etal proposed a design that uses both color camera(s) and range sensor(s) to achieve accurate localization~\cite{evChargingLong2019}. Lou \etal proposed a cable-driven auto-charging robot that uses a vision sensor to localize charging ports~\cite{evChargingLou2020}. However, the experiments were limited to a lab environment and used a simulated charging port. For both~\cite{evChargingLong2019} and~\cite{evChargingLou2020}, localization algorithms were not specifically reported. 

Another relevant work is charging inlet recognition. Sun \etal presented a CNN-based method that predicts whether an input image contains a complete, incomplete, fake charging port, or none~\cite{evChargingSun2018}. However, since it does not include localization, it is not enough for automatic charging robots.

To the best of our knowledge, regarding charging inlet detection, previous works only conducted experiments using a simulated charging inlet under a lab environment. Also, the dataset is currently not publicly available. Hence, in this paper, we collected a novel dataset from real vehicles and will make it publicly available for further research.

%%%%%%%%%%%%%%%%%%%%%%%%%%%%%%%%%%%%%%%%%%%%%%%%%%%%%%%%%%%%%%%%%%%%%%%%%
\subsection{Data Augmentation using Day-to-night Image Translation}
Recently, day-to-night image translation has been studied for data augmentation~\cite{augganECCV2018, augganTITS2021, augganAAAI2020, nightDALeeAccess2020, forkGANECCV2020}. Huang \etal proposed AugGAN that focuses on preserving objects on images during translation~\cite{augganECCV2018, augganTITS2021}. They utilized the network to augment training data by day-to-night image translation to improve nighttime vehicle detection. Lin \etal extended AugGAN to Multimodal AugGAN to generate diversely translated images given an input image~\cite{augganAAAI2020}. To achieve this, they utilized a low-dimensional latent vector to specify a condition for a translated image. Lee \etal presented a framework that uses CycleGAN~\cite{cycleGAN2017} to translate daytime images to nighttime ones and trains a vehicle detector using both original and synthetic images~\cite{nightDALeeAccess2020}. Zheng \etal proposed ForkGAN that decouples domain-invariant content and domain-specific style information to translate nighttime images to daytime images or vice versa~\cite{forkGANECCV2020}. It was utilized to improve image localization/retrieval, semantic segmentation, and object detection. 

While it was not utilized for data augmentation, HiDT network is also about day-to-night image translation~\cite{dayToNightAnokhinCVPR2020}. Anokhin \etal presented the network to re-render an image to another one with different illuminations. 

Another related work is translating nighttime images to daytime images to directly utilize the translated images for further applications~\cite{nightToDayAshaICRA2019, forkGANECCV2020}. 
As the method aims to translate a test image to another one in an easier domain rather than augmenting training data, it does not require re-training final networks (e.g., object detection network). However, it requires translating images in a test phase, and accordingly increases computational demands during testing. Hence, it is less appropriate for real-time applications.

The proposed network is related to day-to-night image translation methods for data augmentation~\cite{augganECCV2018, augganTITS2021, augganAAAI2020, nightDALeeAccess2020, forkGANECCV2020}. However, the proposed network utilizes an explainable environment vector to guide translated images. Also, by varying the vector, the network is able to generate various images given an image. 

%%%%%%%%%%%%%%%%%%%%%%%%%%%%%%%%%%%%%%%%%%%%%%%%%%%%%%%%%%%%%%%%%%%%%%%%%
\subsection{Object detection}
The methods for object detection can be divided into two main categories: one-stage methods and multi-stage methods. One-stage methods have advantages in being more efficient and simpler while multi-stage methods are more accurate and more flexible in general. 

R-CNN is one of the earliest object detection methods that use CNNs~\cite{RCNN2014}. 
Its improved versions were also proposed that reduce training and testing time~\cite{fastRCNN2015, fasterRCNN2015, fasterRCNN2017}. 
They are all multi-stage methods that consist of region proposal step and detection step. 

YOLO~\cite{yolo2016} and SSD~\cite{ssd2016} are two of the earliest one-stage methods. These methods employ a single neural network to predict class probabilities and bounding boxes directly from a whole image instead of each proposed region. As they do not have a separate region proposal step, they are usually faster than multi-stage methods. 
While they are faster, their accuracies are often lower. Lin \etal thought that this could be because of the extreme foreground-background class imbalance~\cite{retinaNet2017}. To overcome this, they proposed RetinaNet that utilizes focal loss. 
Tan \etal proposed EfficientDet~\cite{efficientDet2020} that uses a compound scaling method similar to EfficientNet~\cite{efficientNet2019} to generate a family of one-stage object detectors. 

DetectoRS~\cite{detectoRS2020} proposed the recursive feature pyramid and applied it to a multi-stage detector, HTC~\cite{HTC2019}. The recursive pyramid adds an additional feedback connection from the top-down path to the bottom-up path in the feature pyramid network. It also utilizes switchable atrous convolution that outputs the weighted summation of the outputs from two convolution paths with different atrous rates. 

Recently, methods that utilize Transformer~\cite{transformer2017} were also proposed~\cite{DETR2020, deformableDETR2021, DETRACT2021}. Carion \etal proposed DETR that first extracts features using CNNs and employs transformer encoder/decoder~\cite{DETR2020}. Given object queries, the decoder outputs the corresponding embeddings based on the encoder's output in parallel. The embeddings are then processed by a shared feed-forward network to obtain detection results. To reduce computational demands of DETR, Zhu \etal proposed Deformable DETR~\cite{deformableDETR2021}, and Zheng \etal proposed Adaptive Clustering Transformer~\cite{DETRACT2021}.
%Zhu \etal proposed Deformable DETR that reduces computational demands of DETR in training~\cite{deformableDETR2021}. Zheng \etal proposed Adaptive Clustering Transformer also to reduce computational costs of DETR~\cite{DETRACT2021}.

As we propose a novel detection dataset, we experimented with various object detection networks~\cite{retinaNet2017, fasterRCNN2015, fasterRCNN2017, detectoRS2020, DETR2020} and report their performances.

\section{Proposed method}
\label{sec:method}

In order to improve the robustness of EV charging inlet detection given limited data, we need a method for data augmentation. A central component of our proposed approach is the image-to-image translation network that synthesizes images that look similar to those captured in another environment (time, lighting, etc.). To be controllable and explainable, we first propose the environment guide vector that consists of human interpretable values. We then augment training data using the environment guide vector and the image-to-image network, which we call the \textbf{En}vironment \textbf{T}ranslation GAN (EnT-GAN). Lastly, we utilize augmented training data to train a state-of-the-art object detection network to achieve robust EV charging inlet detection. 

In this section, we first introduce a novel EV Charging Inlet (EVCI) dataset in~\sref{sec:dataset}. We then present environment guide vector and EnT-GAN in~\sref{sec:environ_vector} and in~\sref{sec:environ_net}, respectively. Finally, we present detection method in~\sref{sec:detection}.

\begin{figure}[!t] \begin{center}
\begin{minipage}{0.4\linewidth}
\centerline{\includegraphics[scale=0.4]{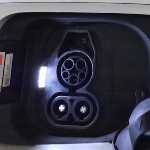}}
\end{minipage}
\begin{minipage}{0.4\linewidth}
\centerline{\includegraphics[scale=0.4]{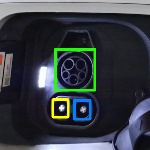}}
\end{minipage}
   \caption{Annotation. Three components are labeled. Left: Cropped image; Right: Cropped image with annotation. Green, yellow, and blue boxes represent top, bottom-left, and bottom-right parts, respectively.}
\label{fig:annotation}
\end{center}\end{figure}

\begin{figure}[!t] \begin{center}
\begin{minipage}{0.24\linewidth}
\centerline{\includegraphics[scale=0.115]{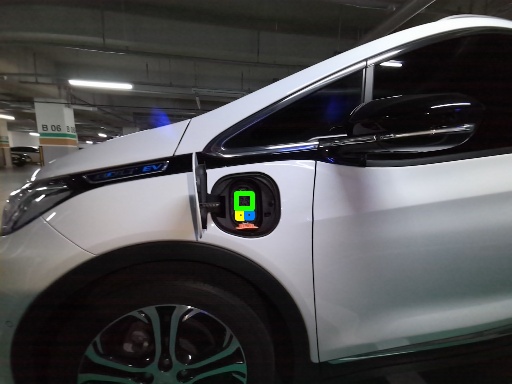}}
\end{minipage}
\begin{minipage}{0.24\linewidth}
\centerline{\includegraphics[scale=0.115]{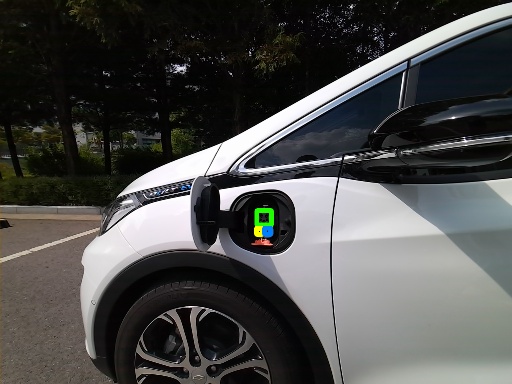}}
\end{minipage}
\begin{minipage}{0.24\linewidth}
\centerline{\includegraphics[scale=0.115]{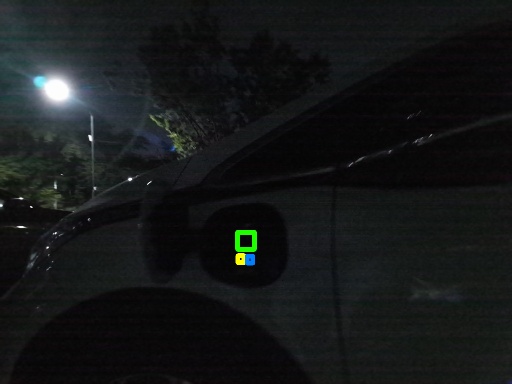}}
\end{minipage}
\begin{minipage}{0.24\linewidth}
\centerline{\includegraphics[scale=0.115]{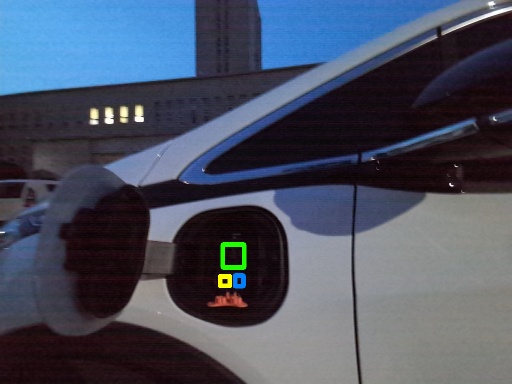}}
\end{minipage}
\\
\vspace{0.05cm}
\begin{minipage}{0.24\linewidth}
\centerline{\footnotesize{(Bolt, Indoor)}}
\end{minipage}
\begin{minipage}{0.24\linewidth}
\centerline{\footnotesize{(Bolt, Daytime)}}
\end{minipage}
\begin{minipage}{0.24\linewidth}
\centerline{\footnotesize{(Bolt, Night)}}
\end{minipage}
\begin{minipage}{0.24\linewidth}
\centerline{\footnotesize{(Bolt, Evening)}}
\end{minipage}
\\
\vspace{0.05cm}
\begin{minipage}{0.24\linewidth}
\centerline{\includegraphics[scale=0.115]{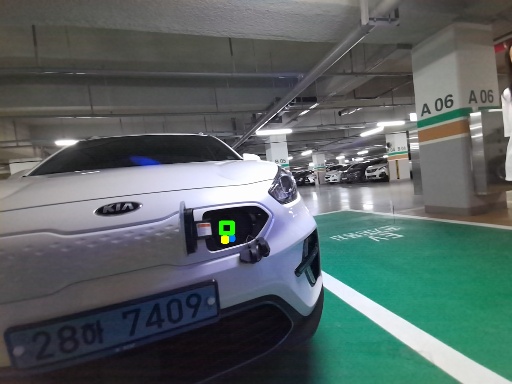}}
\end{minipage}
\begin{minipage}{0.24\linewidth}
\centerline{\includegraphics[scale=0.115]{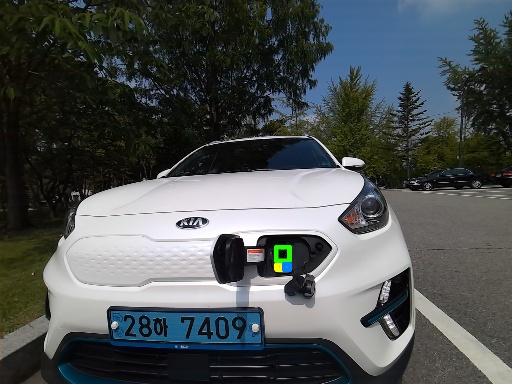}}
\end{minipage}
\begin{minipage}{0.24\linewidth}
\centerline{\includegraphics[scale=0.115]{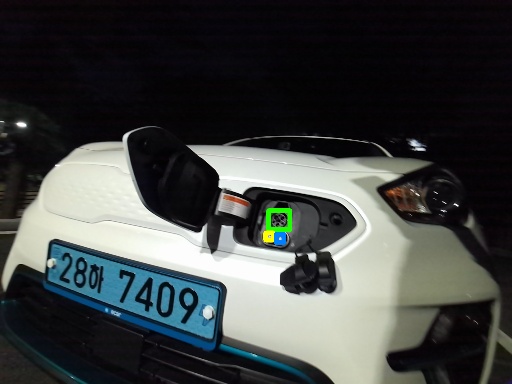}}
\end{minipage}
\begin{minipage}{0.24\linewidth}
\centerline{\includegraphics[scale=0.115]{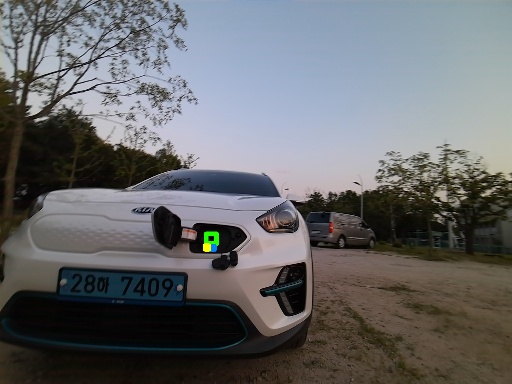}}
\end{minipage}
\\
\vspace{0.05cm}
\begin{minipage}{0.24\linewidth}
\centerline{\footnotesize{(Niro, Indoor)}}
\end{minipage}
\begin{minipage}{0.24\linewidth}
\centerline{\footnotesize{(Niro, Morning)}}
\end{minipage}
\begin{minipage}{0.24\linewidth}
\centerline{\footnotesize{(Niro, Night)}}
\end{minipage}
\begin{minipage}{0.24\linewidth}
\centerline{\footnotesize{(Niro, Evening)}}
\end{minipage}
   \caption{Example images in EVCI Dataset.}
\label{fig:dataset}
\end{center}\end{figure}

\subsection{EV Charging Inlet (EVCI) dataset}
\label{sec:dataset}
We collected a new dataset using Intel RealSense Depth Camera D435i since we were not able to find a publicly available dataset for EV inlet detection. The collected dataset consists of 4,153 pairs of color images and depth maps. Depth maps were only used to get more robust ground truth annotations, especially for images collected at night. The dataset was collected from two vehicles at various environments (locations, times, and others). The two vehicles are Chevrolet Bolt EV 2019 and Kia Niro EV 2018. Ground truth bounding boxes were labeled by human annotators. Specifically, three components of a charging inlet were annotated so that not only location but also rotation can also be estimated (see~\fref{fig:annotation}). Example images are shown in~\fref{fig:dataset}, and detailed information is provided in~\tref{tab:dataset}.

Among 4,153 images, 1,009 and 1,194 images were collected at daytime/morning and at night/evening, respectively. The remaining 1,950 images were collected at indoor parking facilities at various times. The dataset is used to create two benchmark datasets (EVCI-A and EVCI-B dataset). EVCI-A dataset is designed to train, validate, and test on all environments (indoor, daytime/morning, and night/evening) (see~\tref{tab:datasetA}). For the EVCI-B dataset, the training dataset consists of images collected in daytime/morning and at indoor facilities while the validation/testing datasets consist of images captured at night/evening and indoor facilities mostly. Hence, by utilizing the EVCI-B dataset, given only images from daytime/morning/indoor for training, performances on images from night/evening can be measured. 

\begin{table}[!t]
\begin{center}
\begin{minipage}{0.95\linewidth}
 \caption{EVCI Dataset.}
\label{tab:dataset}
\begin{tabu}{X[c,m]|X[c,m]|X[c,m]|X[c,m]|c} 
\hline
Vehicle & Places & Time & Weather & \# of images \\
\hline\hline
        & Indoor  & - & - &  1,059 \\
\cline{2-5}
\multirow{2}{*}{Bolt} &       & Daytime & Sunny & 364 \\
\multirow{2}{*}{EV} & \multirow{2}{*}{Outdoor} & Morning & Rainy & 341 \\
        &         & Night   & Sunny & 326 \\
        &         & Evening  & Sunny & 373 \\
\hline
        & Indoor  & - & - & 891 \\
\cline{2-5}
\multirow{2}{*}{Niro} &       & Daytime & Sunny & 151 \\
\multirow{2}{*}{EV} & \multirow{2}{*}{Outdoor} & Morning & Sunny & 153 \\
	 	&         & Night   & Sunny & 254 \\
        &         & Evening  & Sunny & 241 \\
\hline
\end{tabu}
\end{minipage}
\end{center}
\end{table}

\begin{table}[!t]
\begin{center}
\begin{minipage}{0.95\linewidth}
\caption{Dataset split for EVCI-A dataset and EVCI-B dataset.}
\label{tab:datasetA}
\begin{tabu}{X[c,m]|X[c,m]|c|X[c,m]|X[c,m]} 
\hline
Dataset & \multirow{2}{*}{Places} & \multirow{2}{*}{Time} & \multicolumn{2}{c}{\# of images} \\
\cline{4-5}
split  &  &  & EVCI-A & EVCI-B \\
\hline\hline
         & Indoor    & -                            & 1,273  & 1,207 \\
\cline{2-5}
Training & \multirow{2}{*}{Outdoor} & Daytime/Morning & 562  & 832\\
         &           & Night/Evening                & 409  & - \\
\hline
            & Indoor    & -                             & 267  & 319 \\
\cline{2-5}
Validation  & \multirow{2}{*}{Outdoor} & Daytime/Morning & 74 & - \\
            &           & Night/Evening                 & 196 & 287 \\
\hline
     & Indoor & -                               & 410 & 424 \\
\cline{2-5}
Test & \multirow{2}{*}{Outdoor} & Daytime/Morning & 373 & 177 \\
     &        & Night/Evening                   & 589 & 907 \\
\hline
\end{tabu}
\end{minipage}
\end{center}
\end{table}

\subsection{Environment Guide Vector}
\label{sec:environ_vector}
In general, contrast, brightness, and saturation vary depending on whether an image is captured under daytime lighting levels or night-time lighting~\cite{imgProcessingToNight2002}. Since our goal is, given an image, generating various images that look like images captured under different lighting conditions, we propose to utilize brightness, contrast, and saturation to guide an environment. So, given an image ($\mI$) and an environment guide vector ($\ve\in\calR^{3}$), a corresponding image is generated. 

\begin{figure}[!t] \begin{center}
\begin{minipage}{0.48\linewidth}
\centerline{\includegraphics[scale=0.27]{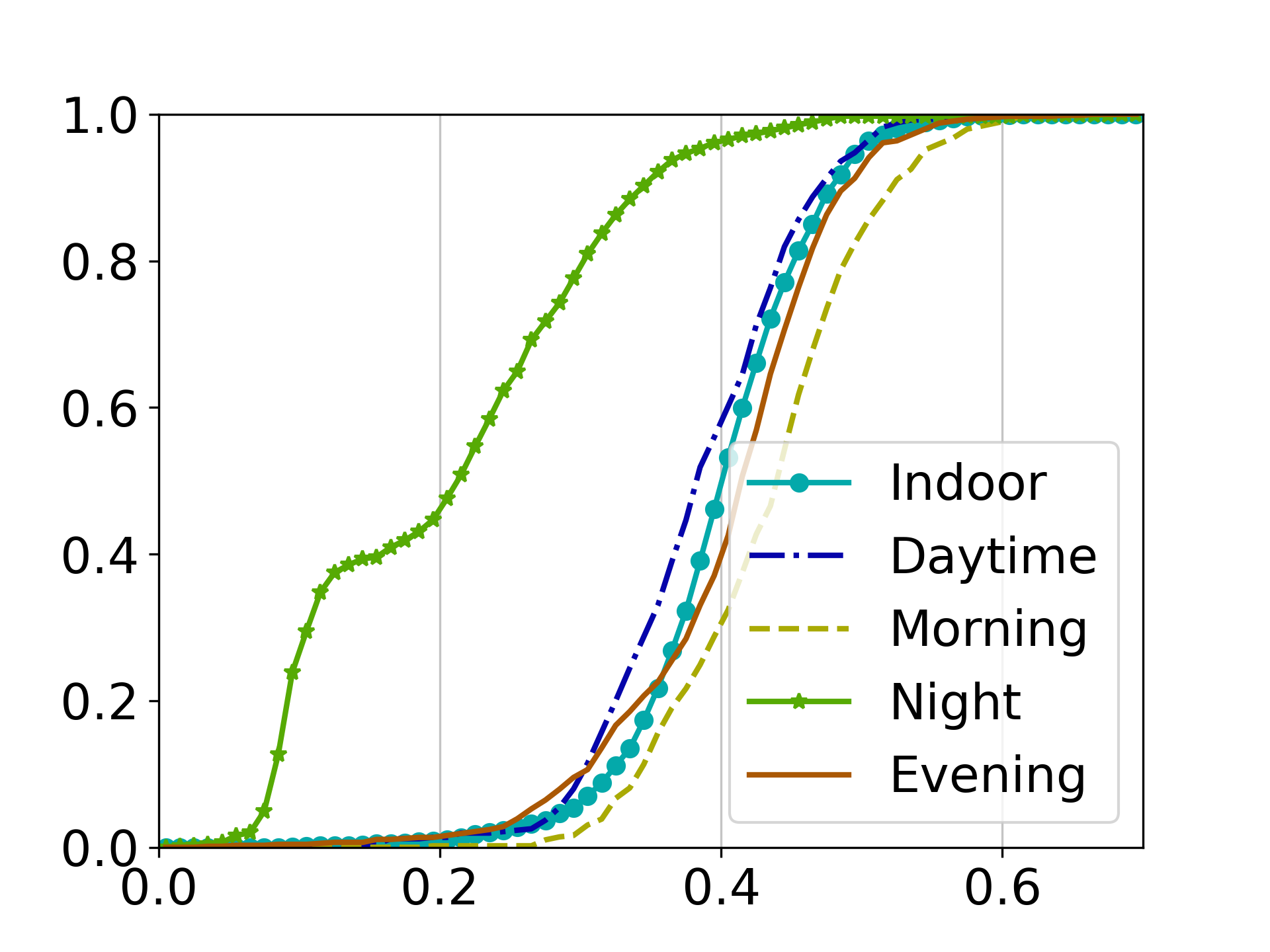}}
\end{minipage}
\begin{minipage}{0.48\linewidth}
\centerline{\includegraphics[scale=0.27]{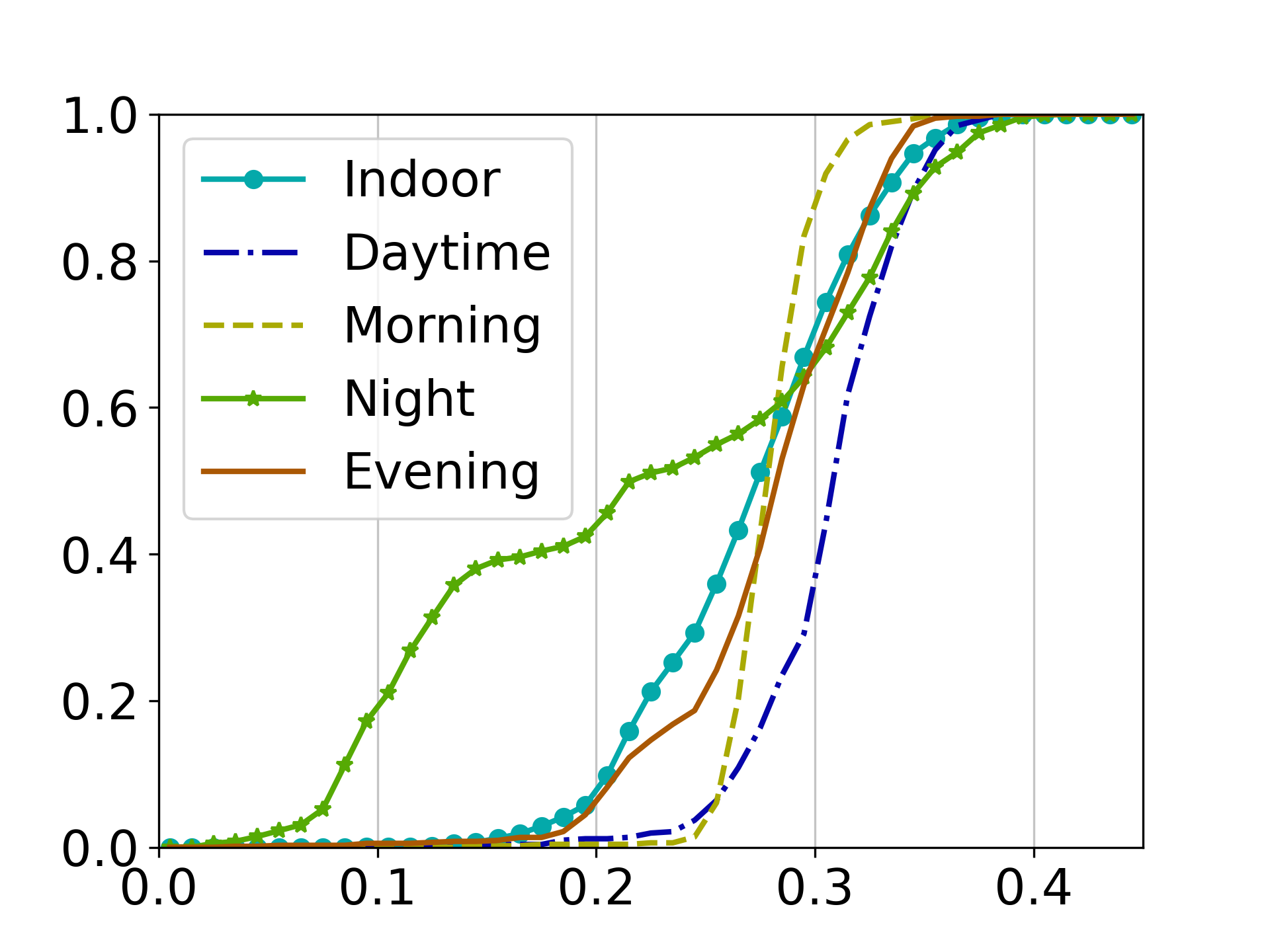}}
\end{minipage}
\\
\begin{minipage}{0.48\linewidth}
\centerline{\footnotesize{(a)}}
\end{minipage}
\begin{minipage}{0.48\linewidth}
\centerline{\footnotesize{(b)}}
\end{minipage}
\\
\begin{minipage}{0.48\linewidth}
\centerline{\includegraphics[scale=0.27]{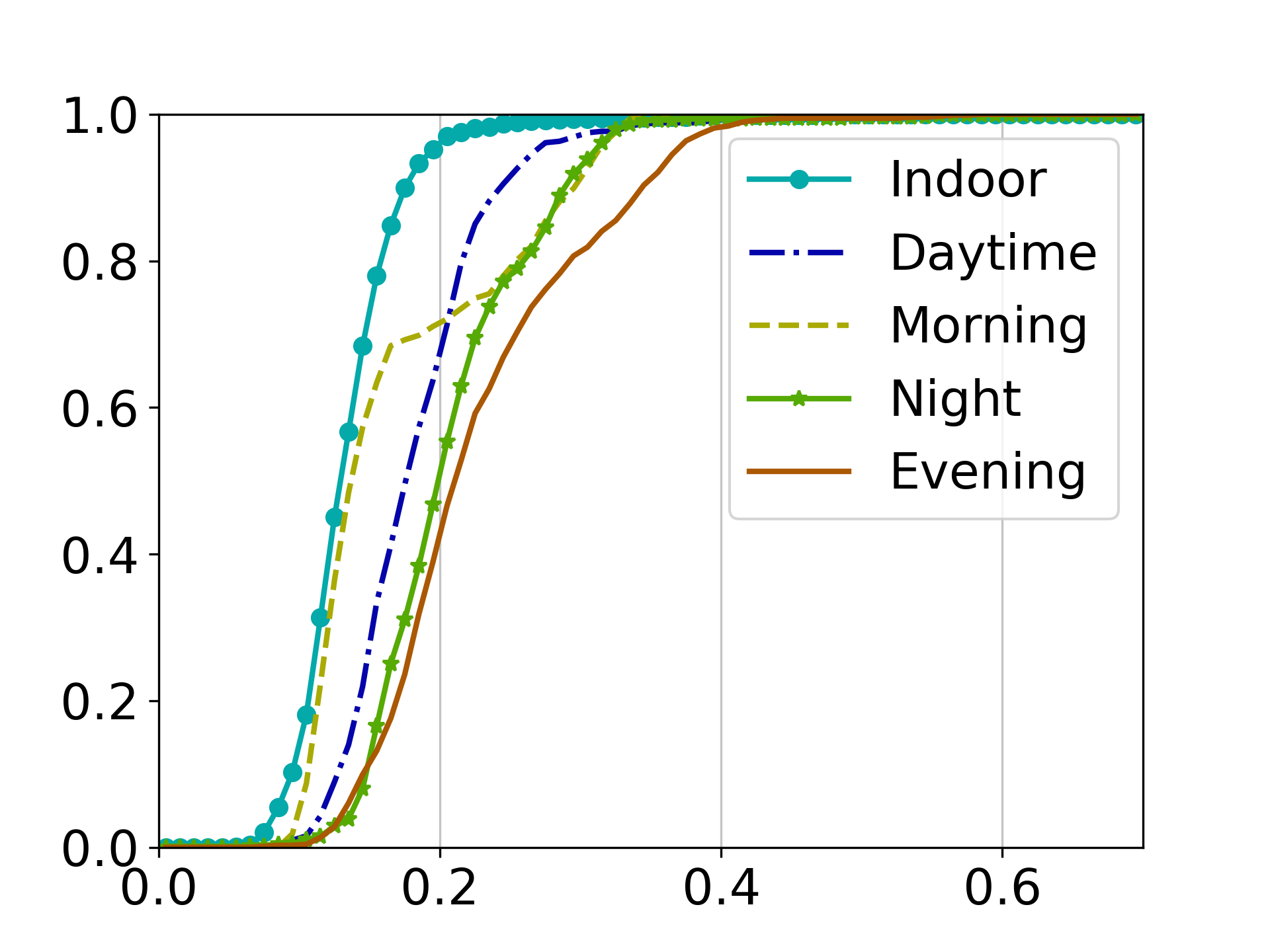}}
\end{minipage}
\\
\begin{minipage}{0.48\linewidth}
\centerline{\footnotesize{(c)}}
\end{minipage}
   \caption{Cumulative distribution of the EVCI dataset. (a) Brightness; (b) RMS contrast; (c) Saturation.}
\label{fig:cum_dist}
\end{center}\end{figure}

\begin{figure}[!t] \begin{center}
\begin{minipage}{0.48\linewidth}
\centerline{\includegraphics[scale=0.12]{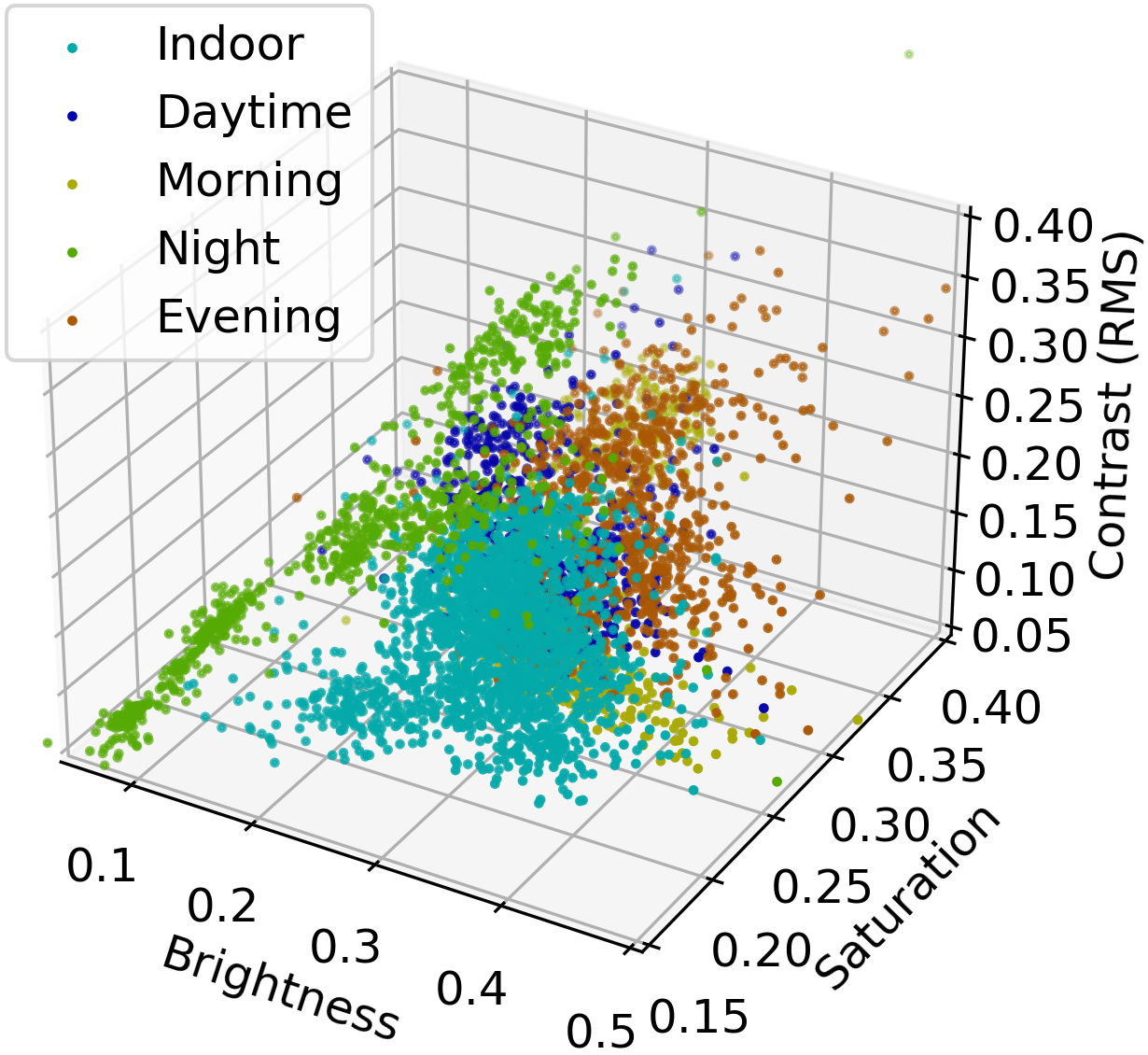}}
\end{minipage}
\begin{minipage}{0.48\linewidth}
\centerline{\includegraphics[scale=0.12]{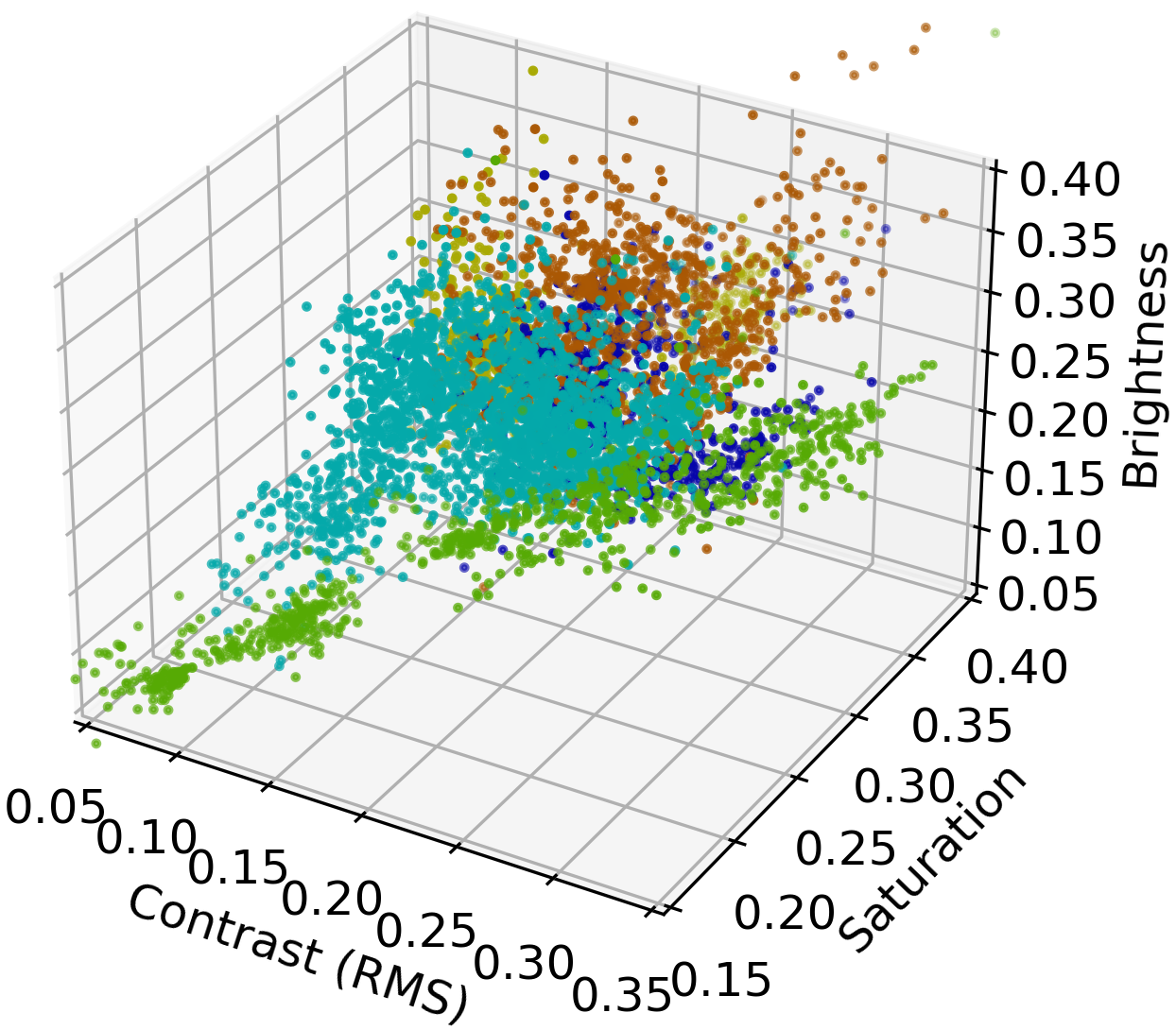}}
\end{minipage}
   \caption{Scatter plot of the EVCI dataset.}
\label{fig:scatter_dist}
\end{center}\end{figure}

To validate the effectiveness of the proposed environment guide vector, we analyzed the brightness, contrast, and saturation of the images in the EVCI dataset. Specifically, cumulative distribution functions of each of them are computed for images captured under indoor, daytime, morning, night, evening environments (see~\fref{fig:cum_dist}). 
We also show the 3D scatter plot of data where each axis corresponds to each of them in~\fref{fig:scatter_dist}. The left and right plots are the same except for their axes.
We can observe that most night-time images have low brightness and low contrast. We can also observe that most indoor images have low saturation while most evening images have high saturation. Please note that the distribution might be different from natural images as the captured environments might have lighting from vehicles, street lamps, or any artificial lighting (see~\fref{fig:annotation} and \fref{fig:dataset}). 

\begin{figure*}[!t] \begin{center}
\begin{minipage}{0.49\linewidth}
\centerline{\includegraphics[scale=0.55]{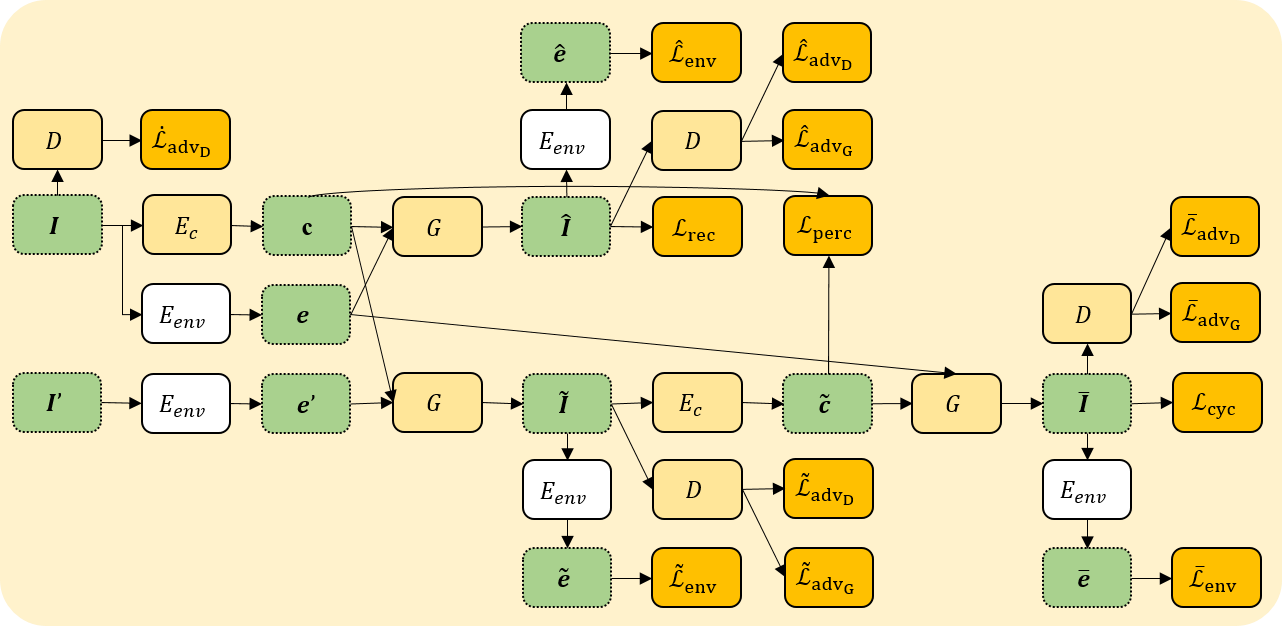}}
\end{minipage}
   \caption{Architecture of the proposed EnT-GAN in training. $\mI$, $\ve$, and $\vc$ represent an image, an environment guide vector, and an encoded content map, respectively. $E_c$, $E_{env}$, $G$, and $D$ denote the content encoder, the environment guide vector extractor, the generator, and the discriminator, respectively. $\calL_{rec}$, $\calL_{cyc}$, $\calL_{env}$, $\calL_{perc}$, and $\calL_{adv}$ represent the image reconstruction loss, the cycle consistency loss, the environment vector consistency loss, the perceptual loss, and the adversarial loss, respectively. Given an image, $E_c$ and $E_{env}$ encode content map $\vc$ and environment guide vector $\ve$, respectively. $G$ synthesizes an image given $\vc$ and $\ve$. $E_{env}$ is a non-learnable component.}
\label{fig:im2im_translation}
\end{center}\end{figure*}

In more detail, brightness ($\ve'_1$) is computed by the mean of pixel-wise luminance, which is calculated by the method in~\cite{rgb2gray}. Root mean square (RMS) contrast is used for contrast ($\ve'_2$), which is the standard deviation of pixel-wise luminance. Saturation ($\ve'_3$) is computed by the mean of pixel-wise saturation, which is calculated by the same way as the HSV color representation. The environment vector $\ve$ is then determined by normalizing each component of the computed vector $\ve'$ to $[-1,1]$. The normalization is processed using the minimum and maximum of each component of all images in the dataset.

\subsection{Environment Translation GAN (EnT-GAN)}
\label{sec:environ_net}

The proposed image-to-image translation network is designed to be able to synthesize an image that a human can easily control and explain. Also, it is to generate various images given an image. The training framework of the proposed network is shown in~\fref{fig:im2im_translation}. $E_c$ and $E_{env}$ represent the content encoder and the environment guide vector extractor, respectively. $E_c$ aims to encode content information and consists of three convolution layers and five residual blocks where each residual block contains two convolution layers and one skip connection. $E_{env}$ aims to extract environment information such as lighting conditions, the existence of artificial lighting, etc. It is designed to be able to intuitively control synthesized images. It extracts brightness, contrast, and saturation as explained in~\sref{sec:environ_vector} and is a non-learnable component. $G$ represents the generator that synthesizes an image given an encoded content map $\vc$ and an environment guide vector $\ve\in\calR^{3}$. The environment vector is first processed by an MLP that consists of three fully connected layers. Then, adaptive instance normalization (AdaIN)~\cite{adain} is applied to the encoded content map using the MLP-processed environment vector. Finally, the output of the AdaIN layer is processed by four residual blocks, two transposed convolution layers, and one convolution layer to obtain synthesized images.

In the inference stage, given an image and an environment vector, the generator $G$ synthesizes an image along with the content encoder $E_c$. The environment vector can either be human selected, randomly generated, or extracted from an image (using environment vector extractor $E_{env}$). 

In the training phase, the network is trained using the following objective functions: 
\begin{equation*}
\begin{split}
    \min_{E_c, G} \calL_{E,G}  \\
    \min_{D} \calL_{D}  
\end{split}
\end{equation*}
where $\calL_{E,G}$ denotes the loss that is used to train the generator and the content encoder. The loss consists of image reconstruction loss, cycle consistency loss, environment translation loss, perceptual loss, and adversarial loss for generators. $\calL_{D}$ contains the adversarial loss for discriminators.

\textbf{Image reconstruction loss.} The loss $\calL_{rec}$ is to be able to reconstruct an input image if its encoded content map $\vc$ and its extracted environment vector $\ve$ are given to the generator $G$ as inputs. 
The loss is defined as the $\ell^1$-norm of the pixel-wise differences between the input image $\mI$ and the reconstructed image $\hat{\mI}$. Formally, it is defined as $\calL_{rec} = \norm{\hat{\mI} - \mI}_1 = \norm{G(\vc, \ve) - \mI}_1 = \norm{G(E_c(\mI), E_{env}(\mI)) - \mI}_1$.

\textbf{Cycle consistency loss.} The loss $\calL_{cyc}$ has similarity with the image reconstruction loss as it is also about reconstructing an input image. The difference is that the reconstructed image $\bar{\mI}$ is generated given the encoded content map $\tilde{\vc}$ of another synthesized image $\tilde{\mI}$ along with the input image's environment vector $\ve$. $\tilde{\mI}$ is generated using the encoded content map $\vc$ of the target input image $\mI$ and an environment vector $\ve'$ from another image $\mI'$. It is defined as follows:
\begin{equation*}
\begin{split}
\calL_{cyc} &= \norm{\bar{\mI} - \mI}_1 = \norm{G(\tilde{\vc}, \ve) - \mI}_1 \\
&= \norm{G(E_c(\tilde{\mI}), E_{env}(\mI)) - \mI}_1  					\\
&= \norm{G(E_c(G(E_c(\mI), \ve')), E_{env}(\mI)) - \mI}_1 				\\
&= \norm{G(E_c(G(E_c(\mI), E_{env}(\mI'))), E_{env}(\mI)) - \mI}_1.  
\end{split}
\end{equation*}

\textbf{Environment translation loss.} The loss $\calL_{env}$ is to ensure that a generated image has the characteristics that are guided by the given environment vector. It measures the differences between the given environment guide vector to a generator and the extracted environment vector from the synthesized image from the generator. 
Considering~\fref{fig:im2im_translation}, as three generators exist, three corresponding environment translation losses are computed. The total environment translation loss $\calL_{env}$ is the summation of them ($\hat{\calL}_{env}$, $\tilde{\calL}_{env}$, $\bar{\calL}_{env}$).
\begin{equation*}
\begin{split}
\calL_{env} =& \hat{\calL}_{env} + \tilde{\calL}_{env} + \bar{\calL}_{env} \\
=& \norm{\hat{\ve} - \ve}_1 + \norm{\tilde{\ve} - \ve'}_1 + \norm{\bar{\ve} - \ve}_1 \\
=& \norm{E_{env}(G(E_c(\mI), E_{env}(\mI))) - E_{env}(\mI)}_1 \\
+& \norm{E_{env}(G(E_c(\mI), E_{env}(\mI'))) - E_{env}(\mI')}_1 \\
+& \norm{E_{env}(G(E_c(G(E_c(\mI), E_{env}(\mI'))), E_{env}(\mI))) \\ 
&- E_{env}(\mI)}_1.
\end{split}
\end{equation*}

\textbf{Perceptual loss.} The loss $\calL_{perc}$ is to enforce the consistency of contents between an input image and the generated image using the encoded content map of the input image. To achieve this, the loss measures the difference between the encoded content map $\vc$ of the input image and that $\tilde{\vc}$ of the synthesized image $\tilde{\mI}$ where the latter image is generated using $\vc$ and another image's environment vector $\ve'$.
The loss is defined as follows:
\begin{equation*}
\begin{split}
\calL_{perc} =& \norm{\tilde{\vc} - \vc}_1 = \norm{E_c(\tilde{\mI}) - E_c(\mI)}_1    \\
 =& \norm{E_c(G(E_c(\mI), \ve')) - E_c(\mI)}_1.
\end{split}
\end{equation*}

\textbf{Adversarial loss.} The loss $\calL_{adv}$ is employed to ensure that the distribution of generated images is similar to that of real images in the dataset. Similar to~\cite{forkGANECCV2020, cycleGAN2017, dayToNightAnokhinCVPR2020}, we utilize the least squares loss from LSGAN~\cite{lsgan}. We also employ multi-scale discriminators where we used two scales. One discriminator is applied at generated images' resolution, and the other takes images at half resolution. It is defined as follows:
\begin{equation*}
\begin{split}
\calL_{adv_G} = & \hat{\calL}_{adv_G} + \tilde{\calL}_{adv_G} + \bar{\calL}_{adv_G} \\ 
			  = (D&(\hat{\mI})-1)^2 + (D(\tilde{\mI})-1)^2 + (D(\bar{\mI})-1)^2     \\
			  + (D&(S(\hat{\mI}))-1)^2 + (D(S(\tilde{\mI}))-1)^2 + (D(S(\bar{\mI}))-1)^2     
\end{split}
\end{equation*}

\begin{equation*}
\begin{split}
\calL_{adv_D} = & \dot{\calL}_{adv_D} + \hat{\calL}_{adv_D} + \tilde{\calL}_{adv_D} + \bar{\calL}_{adv_D} \\ 
			  = (D(&\mI) -1)^2 + (D(\hat{\mI})-0)^2 + (D(\tilde{\mI})-0)^2 \\
                + (D(&\bar{\mI})-0)^2 +(D(S(\mI))-1)^2 + (D(S(\hat{\mI}))-0)^2 \\
                + (D(&S(\tilde{\mI}))-0)^2 + (D(S(\bar{\mI})) -0)^2    
\end{split}
\end{equation*}
where $S(\cdot)$ denotes downsampling by 2. 

\textbf{Final loss.} The generator and the content encoder are trained by minimizing $\calL_{E,G}$ that is the weighted summation of $\calL_{rec}$, $\calL_{cyc}$, $\calL_{env}$, $\calL_{perc}$, and $\calL_{adv_G}$. The discriminator is trained by minimizing $\calL_{D}$.
\begin{equation*}
\begin{split}
&     \calL_{E,G} = \lambda_1 \calL_{rec} + \lambda_2 \calL_{cyc} + \lambda_3 \calL_{env} + \lambda_4 \calL_{perc} + \lambda_5 \calL_{adv_G} \\ 
&    \calL_{D} = \lambda_6 \calL_{adv_D} 
\end{split}
\end{equation*}

The network is trained using the Adam optimizer for 20 epochs with the initial learning rate of 0.0002. The learning rate was linearly decayed after 10 epochs. We used the following hyperparameters: $\lambda_1 = \lambda_2 = 10/hw$, $\lambda_3 = 0.5$, $\lambda_4 = 1/h'w'$, $\lambda_5 = \lambda_6 = 0.5$ where $(h, w)$ and $(h', w')$ denote the height and width of an input image and the corresponding content map, respectively. In the training phase, the input image is resized to $286\times572$ and then is scaled by a factor that is randomly sampled from the uniform distribution $[0.9, 1.1]$. It is then randomly cropped to be $256\times512$ and is processed by random horizontal flipping.

\subsection{Detection}
\label{sec:detection}

To demonstrate that the data augmentation using the proposed network improves a state-of-the-art detection network, we first analyzed multiple detection networks~\cite{fasterRCNN2015, fasterRCNN2017, retinaNet2017, detectoRS2020, DETR2020} (see~\tref{tab:result1}). Among them, the highest accuracy is achieved by DetectoRS~\cite{detectoRS2020}. Hence, we compare the performances of it trained using the original training dataset, using the ForkGAN-augmented dataset~\cite{forkGANECCV2020}, and using the proposed EnT-GAN-augmented dataset.

For both ForkGAN~\cite{forkGANECCV2020} and the proposed EnT-GAN, a detection network is trained using augmented training data that consists of both original training images and translated images. The distribution of data during training is 50\% from original training images and the remaining 50\% from translated images. For the proposed EnT-GAN, environment vectors are randomly generated from the uniform distribution $[-1, 1]$ unless specifically mentioned. Also, since EnT-GAN can generate various images given an image, varying sets of translated images are used during training.  

The detection network is trained using stochastic gradient descent (SGD) optimizer for 30 epochs with linearly increasing learning rate until 500 iterations to 0.002/0.001 for the EVCI-A/B dataset. The input image is first resized to $800 \times 1333$ while keeping aspect ratio and is randomly flipped horizontally. We also show the result of utilizing random scaling/cropping. For the random scaling, a scaling factor is sampled from the uniform distribution $[0.9, 1.1]$. The scaled image is then randomly cropped to be $384 \times 600$.

% The detection network is trained using stochastic gradient descent (SGD) optimizer for 30 epochs with linearly increasing learning rate until 500 iterations to 0.002 for the EVCI-A dataset and 0.001 for the EVCI-B dataset. The input image is first resized to $800 \times 1333$ while keeping aspect ratio and is randomly flipped horizontally. We also show the result of utilizing random scaling/cropping. For the random scaling, a scaling factor is sampled from the uniform distribution $[0.9, 1.1]$. The scaled image is then randomly cropped to be $384 \times 600$.

\begin{figure*}[!t] \begin{center}
\begin{minipage}{0.17\linewidth}
\centerline{\includegraphics[scale=0.12]{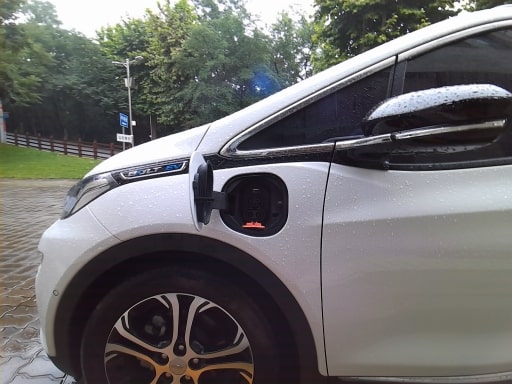}}
\end{minipage}
\begin{minipage}{0.17\linewidth}
\centerline{\includegraphics[scale=0.12]{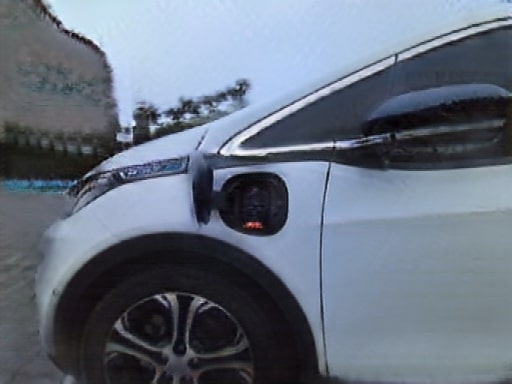}}
\end{minipage}
\begin{minipage}{0.17\linewidth}
\centerline{\includegraphics[scale=0.12]{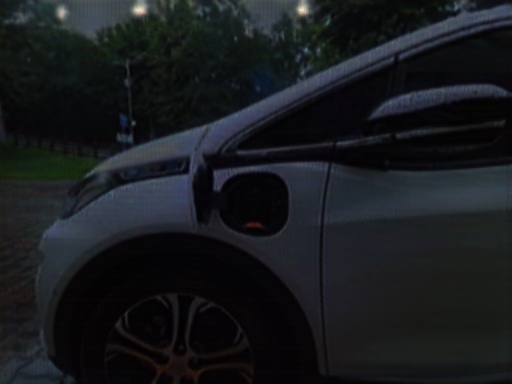}}
\end{minipage}
\begin{minipage}{0.17\linewidth}
\centerline{\includegraphics[scale=0.12]{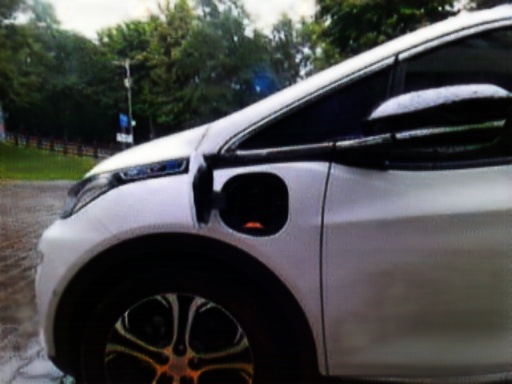}}
\end{minipage}
\begin{minipage}{0.17\linewidth}
\centerline{\includegraphics[scale=0.12]{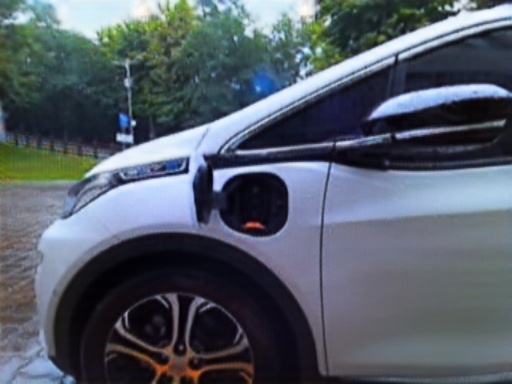}}
\end{minipage}
\\
\vspace{0.05cm}
\begin{minipage}{0.17\linewidth}
\centerline{\footnotesize{Original image}}
\end{minipage}
\begin{minipage}{0.17\linewidth}
\centerline{\footnotesize{ForkGAN~\cite{forkGANECCV2020}}}
\end{minipage}
\begin{minipage}{0.17\linewidth}
\centerline{\footnotesize{$\ve=(-0.8, -0.8, -0.8)$}}
\end{minipage}
\begin{minipage}{0.17\linewidth}
\centerline{\footnotesize{$\ve=(-0.3, 0.5, -0.2)$}}
\end{minipage}
\begin{minipage}{0.17\linewidth}
\centerline{\footnotesize{$\ve=(0.1, 0.1, 0.1)$}}
\end{minipage}
\\
\vspace{0.05cm}
\begin{minipage}{0.17\linewidth}
\centerline{\includegraphics[scale=0.12]{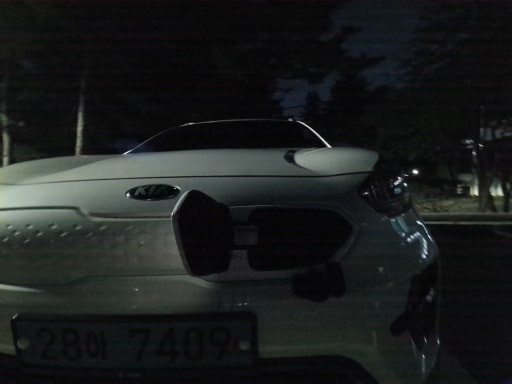}}
\end{minipage}
\begin{minipage}{0.17\linewidth}
\centerline{\includegraphics[scale=0.12]{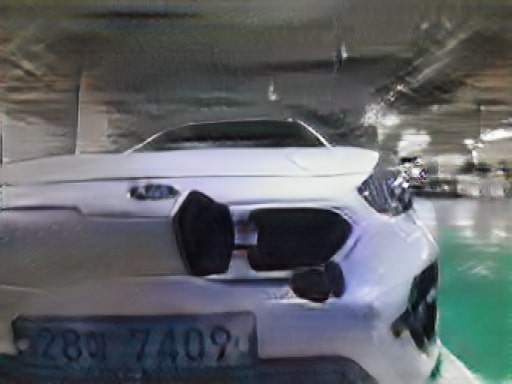}}
\end{minipage}
\begin{minipage}{0.17\linewidth}
\centerline{\includegraphics[scale=0.12]{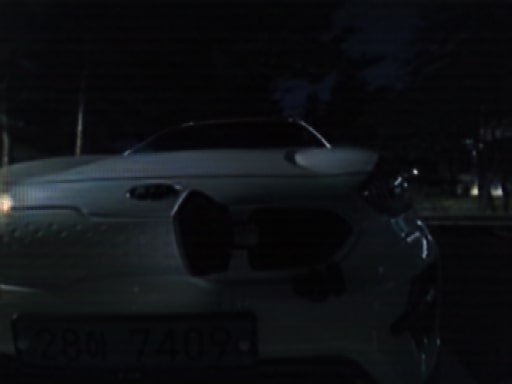}}
\end{minipage}
\begin{minipage}{0.17\linewidth}
\centerline{\includegraphics[scale=0.12]{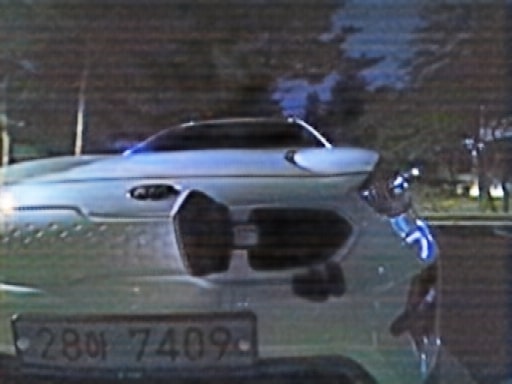}}
\end{minipage}
\begin{minipage}{0.17\linewidth}
\centerline{\includegraphics[scale=0.12]{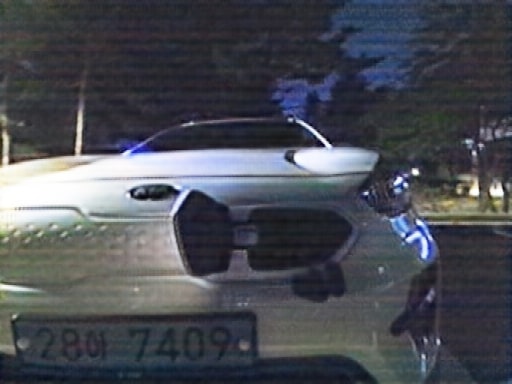}}
\end{minipage}
\\
\vspace{0.05cm}
\begin{minipage}{0.17\linewidth}
\centerline{\footnotesize{Original image}}
\end{minipage}
\begin{minipage}{0.17\linewidth}
\centerline{\footnotesize{ForkGAN~\cite{forkGANECCV2020}}}
\end{minipage}
\begin{minipage}{0.17\linewidth}
\centerline{\footnotesize{$\ve=(-0.8, -0.8, -0.8)$}}
\end{minipage}
\begin{minipage}{0.17\linewidth}
\centerline{\footnotesize{$\ve=(0.3, -0.5, 0.2)$}}
\end{minipage}
\begin{minipage}{0.17\linewidth}
\centerline{\footnotesize{$\ve=(0.8, 0.8, 0.8)$}}
\end{minipage}
\\
\vspace{0.05cm}
\begin{minipage}{0.17\linewidth}
\centerline{\includegraphics[scale=0.12]{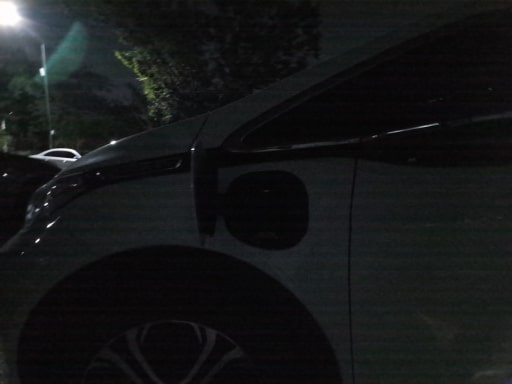}}
\end{minipage}
\begin{minipage}{0.17\linewidth}
\centerline{\includegraphics[scale=0.12]{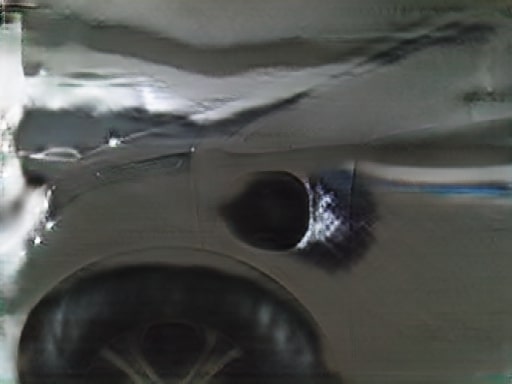}}
\end{minipage}
\begin{minipage}{0.17\linewidth}
\centerline{\includegraphics[scale=0.12]{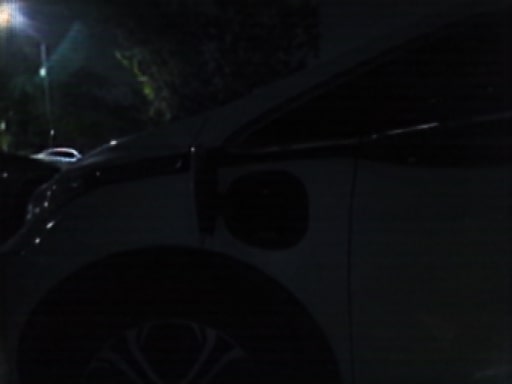}}
\end{minipage}
\begin{minipage}{0.17\linewidth}
\centerline{\includegraphics[scale=0.12]{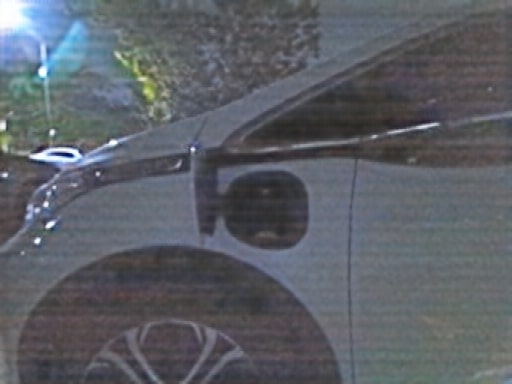}}
\end{minipage}
\begin{minipage}{0.17\linewidth}
\centerline{\includegraphics[scale=0.12]{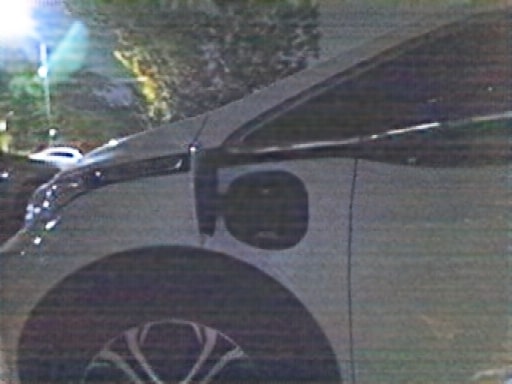}}
\end{minipage}
\\
\vspace{0.05cm}
\begin{minipage}{0.17\linewidth}
\centerline{\footnotesize{Original image}}
\end{minipage}
\begin{minipage}{0.17\linewidth}
\centerline{\footnotesize{ForkGAN~\cite{forkGANECCV2020}}}
\end{minipage}
\begin{minipage}{0.17\linewidth}
\centerline{\footnotesize{$\ve=(-0.8, -0.8, -0.8)$}}
\end{minipage}
\begin{minipage}{0.17\linewidth}
\centerline{\footnotesize{$\ve=(0.3, -0.5, 0.2)$}}
\end{minipage}
\begin{minipage}{0.17\linewidth}
\centerline{\footnotesize{$\ve=(0.8, 0.8, 0.8)$}}
\end{minipage}
   \caption{Qualitative results of image-to-image translation.}
\label{fig:result_im2im}
\end{center}\end{figure*}

\section{Experiments and results}
\label{sec:result}

In this section, we demonstrate the controllability and explainability of the proposed EnT-GAN and the effectiveness of the proposed framework.

\begin{table*}[!t]
\begin{center}
\begin{minipage}{0.9\linewidth}
\caption{Results on EVCI-A dataset and EVCI-B dataset. $AP$ is computed for IoU from 0.5 to 0.95.}
\label{tab:result1}
\begin{tabu}{c|X[c,m]|X[c,m]|c|X[c,m]|X[c,m]} 
\hline
\multicolumn{4}{c|}{Method}  & EVCI-A & EVCI-B \\
\hline
Detection networks & Backbone & EnT-GAN & Random scaling/cropping  & $AP$ & $AP$ \\
\hline\hline
Faster R-CNN							& \multirow{6}{*}{ResNet-101} & -   & - & 59.2 & 54.3  \\ 
\cite{fasterRCNN2015, fasterRCNN2017}	& 							  & \checkmark   & - & 61.6 & 56.5   \\ 
\cline{1-1}\cline{3-6}
RetinaNet    				& & -        & - 				& 58.5  & 55.5  \\   
\cite{retinaNet2017}    	& & \checkmark  & - 			& 59.8  & 57.0  \\   
\cline{1-1}\cline{3-6}
DETR	         		& & - & \checkmark 					& 54.9  & 51.7  \\ 
\cite{DETR2020}		 & & \checkmark & \checkmark 			& 55.7  & 52.4  \\ 
\hline
   					 & \multirow{4}{*}{ResNet-50} & - & - 	& 64.1  & 61.5  \\  
DetectoRS 			 & & ForkGAN~\cite{forkGANECCV2020} & - & 63.6  & 60.4  \\  
\cite{detectoRS2020} & & \checkmark  & - 					& \textbf{65.0} &  \textbf{62.0}  \\
 					 & & \checkmark  & \checkmark 			& \textbf{66.5}  & \textbf{63.0}  \\
\hline
\end{tabu}
\end{minipage}
\end{center}
\end{table*}

\begin{table}[!t]
\begin{center}
\begin{minipage}{0.98\linewidth}
\caption{Ablation Study on Sampling of Environment Guide Vector using DetectoRS~\cite{detectoRS2020} and EVCI-B dataset.}
\label{tab:ablation}
\begin{tabu}{c|X[c,m]} 
\hline
Sampling distribution  & $AP$ \\
\hline\hline
	$U[-1,+1]$ & 62.0 \\  
    Vector from target domain + $U[-0.2,+0.2]$  & \textbf{62.5}  \\  
\hline
\end{tabu}
\end{minipage}
\end{center}
\end{table}

\begin{table}[!t]
\begin{center}
\begin{minipage}{0.98\linewidth}
 \caption{Study on EnT-GAN-based Mosaic Data Augmentation using DetectoRS~\cite{detectoRS2020} and EVCI-B dataset.}
\label{tab:result_mosaic}
\begin{tabu}{X[c,m]|c|X[c,m]} 
\hline
 Mosaic & Random scaling/cropping  & $AP$ \\
\hline\hline
- & - 						& 62.0 \\  
\checkmark & - 			& 62.6 \\  
- & \checkmark 			& 63.0 \\  
\checkmark & \checkmark & \textbf{63.2} \\  
\hline
\end{tabu}
\end{minipage}
\end{center}
\end{table}

\begin{figure}[!t] \begin{center}
\begin{minipage}{0.31\linewidth}
\centerline{\includegraphics[scale=0.17]{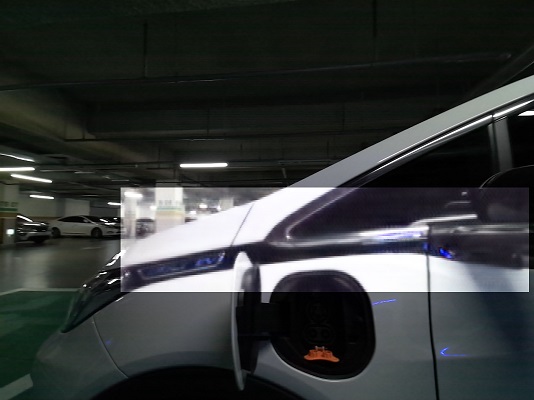}}
\end{minipage}
\begin{minipage}{0.31\linewidth}
\centerline{\includegraphics[scale=0.17]{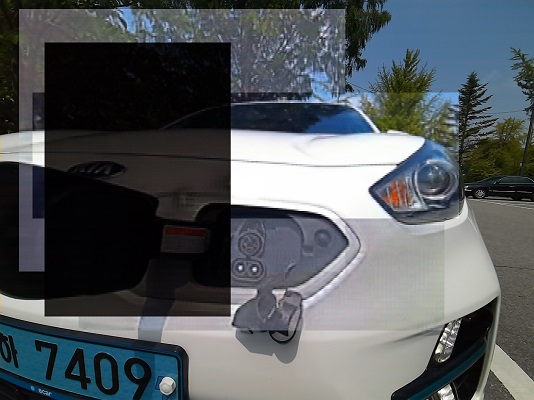}}
\end{minipage}
\begin{minipage}{0.31\linewidth}
\centerline{\includegraphics[scale=0.17]{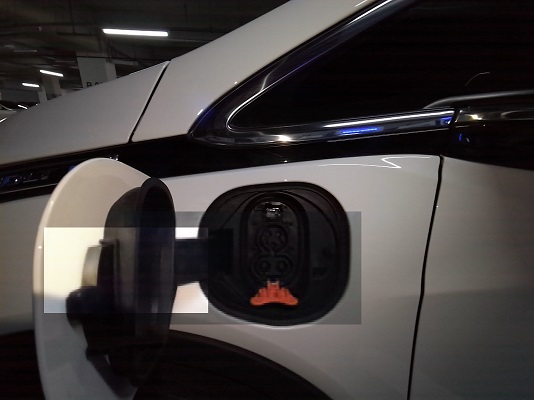}}
\end{minipage}
   \caption{Examples of EnT-GAN-based Mosaic augmentation.}
\label{fig:result_mosaic}
\end{center}\end{figure}

\begin{figure}[!t] \begin{center}
\begin{minipage}{0.44\linewidth}
\centerline{\includegraphics[scale=0.25]{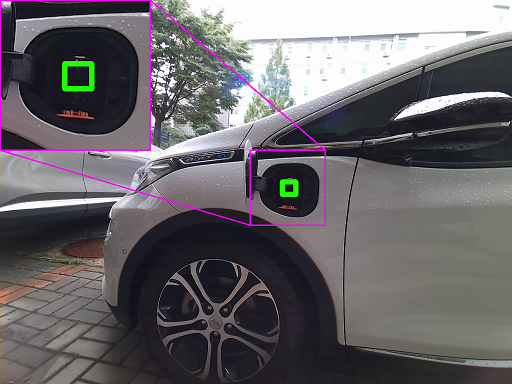}}
\end{minipage}
\begin{minipage}{0.44\linewidth}
\centerline{\includegraphics[scale=0.25]{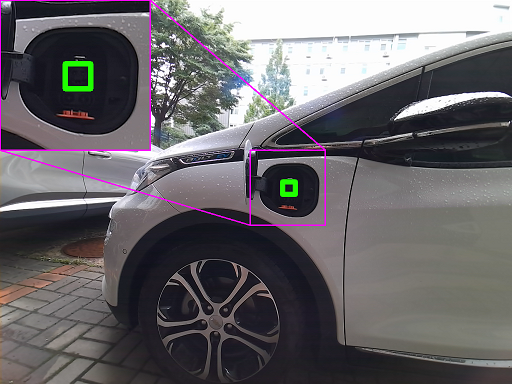}}
\end{minipage}
\\
\vspace{0.05cm}
\begin{minipage}{0.44\linewidth}
\centerline{\footnotesize{DetectoRS~\cite{detectoRS2020}}}
\end{minipage}
\begin{minipage}{0.44\linewidth}
\centerline{\footnotesize{DetectoRS with ForkGAN~\cite{forkGANECCV2020}}}
\end{minipage}
\\
\vspace{0.05cm}
\begin{minipage}{0.44\linewidth}
\centerline{\includegraphics[scale=0.25]{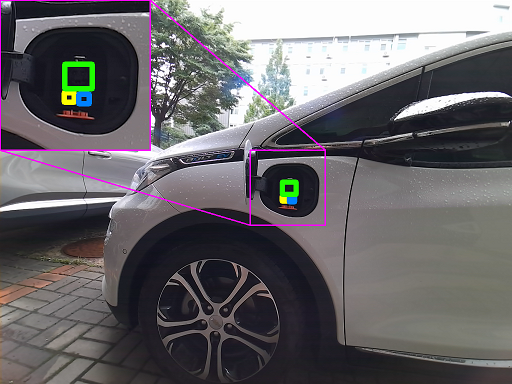}}
\end{minipage}
\begin{minipage}{0.44\linewidth}
\centerline{\includegraphics[scale=0.25]{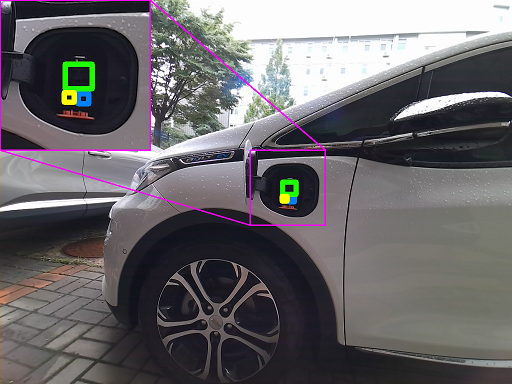}}
\end{minipage}
\\
\vspace{0.05cm}
\begin{minipage}{0.44\linewidth}
\centerline{\footnotesize{Proposed}}
\end{minipage}
\begin{minipage}{0.44\linewidth}
\centerline{\footnotesize{Ground truth}}
\end{minipage}
\\
\vspace{0.05cm}
\begin{minipage}{0.44\linewidth}
\centerline{\includegraphics[scale=0.25]{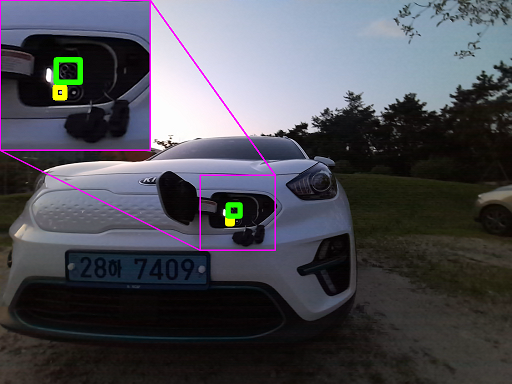}}
\end{minipage}
\begin{minipage}{0.44\linewidth}
\centerline{\includegraphics[scale=0.25]{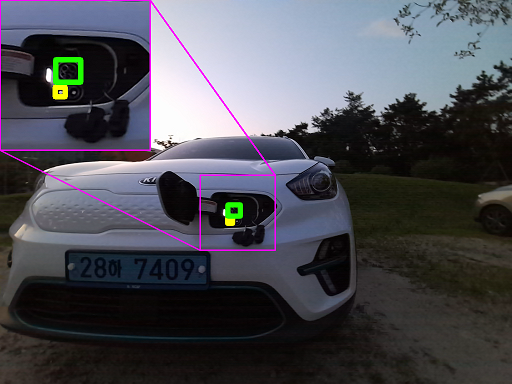}}
\end{minipage}
\\
\vspace{0.05cm}
\begin{minipage}{0.44\linewidth}
\centerline{\footnotesize{DetectoRS~\cite{detectoRS2020}}}
\end{minipage}
\begin{minipage}{0.44\linewidth}
\centerline{\footnotesize{DetectoRS with ForkGAN~\cite{forkGANECCV2020}}}
\end{minipage}
\\
\vspace{0.05cm}
\begin{minipage}{0.44\linewidth}
\centerline{\includegraphics[scale=0.25]{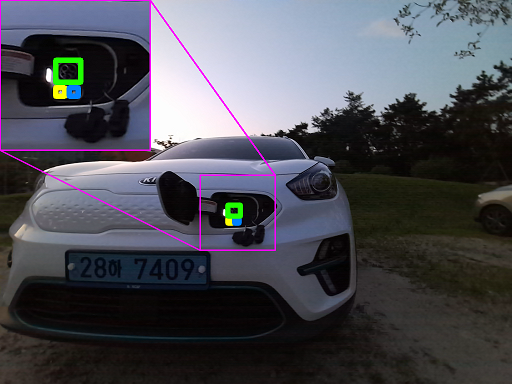}}
\end{minipage}
\begin{minipage}{0.44\linewidth}
\centerline{\includegraphics[scale=0.25]{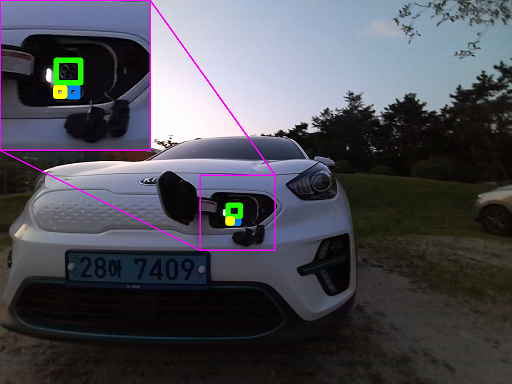}}
\end{minipage}
\\
\vspace{0.05cm}
\begin{minipage}{0.44\linewidth}
\centerline{\footnotesize{Proposed}}
\end{minipage}
\begin{minipage}{0.44\linewidth}
\centerline{\footnotesize{Ground truth}}
\end{minipage}
   \caption{Qualitative comparison of EV charging inlet detection.} 
\label{fig:result}
\end{center}\end{figure}

\begin{figure}[!t] \begin{center}
\begin{minipage}{0.44\linewidth}
\centerline{\includegraphics[scale=0.17]{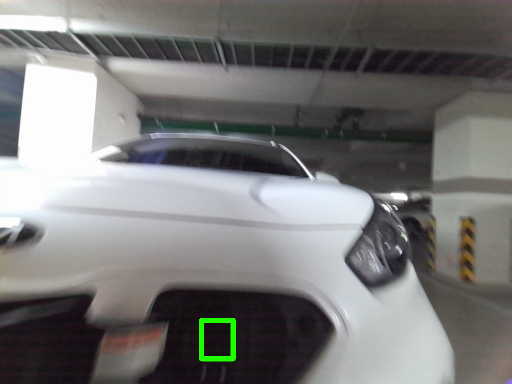}}
\end{minipage}
\begin{minipage}{0.44\linewidth}
\centerline{\includegraphics[scale=0.17]{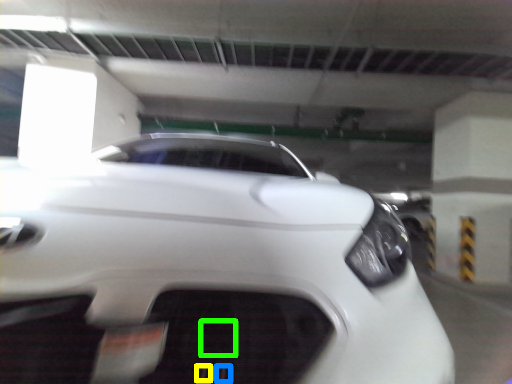}}
\end{minipage}
\\
\vspace{0.05cm}
\begin{minipage}{0.44\linewidth}
\centerline{\footnotesize{Proposed}}
\end{minipage}
\begin{minipage}{0.44\linewidth}
\centerline{\footnotesize{Ground truth}}
\end{minipage}
   \caption{Failure case of the proposed method.}
\label{fig:failure}
\end{center}\end{figure}

\textbf{Image-to-Image Translation.} 
To demonstrate the controllability and the explainability of the proposed EnT-GAN, we show the synthesized images given an image and various environment guide vectors in~\fref{fig:result_im2im} and~\fref{fig:teaser}. This obviously shows that the proposed EnT-GAN can generate various images given an input image by varying environment guide vectors. The left-most column shows the input images, and the second column shows the synthesized images using ForkGAN~\cite{forkGANECCV2020}. The other columns show the synthesized images using the proposed EnT-GAN given various environment vectors (presented at the bottom of each image). As ForkGAN uses predefined two domains, it is only able to translate an image in one domain to the other domain. Consequently, it cannot synthesize various images and is not able to control generated images. The generated images further show that the quality of the EnT-GAN is superior to that of the previous state-of-the-art method~\cite{forkGANECCV2020}.

\textbf{EV Charging Inlet Detection.}
To evaluate EV charging inlet detection quantitatively, mean average precision (mAP) is computed. We report $AP$ that is the average AP for intersection over union (IoU) from 0.5 to 0.95 with a step size of 0.05, which is used in the COCO dataset~\cite{lin2014microsoft}. 

\tref{tab:result1} shows the quantitative results of the proposed method and other comparing methods~\cite{fasterRCNN2015, fasterRCNN2017, detectoRS2020, DETR2020, forkGANECCV2020}. First of all, the experimental results demonstrate that by utilizing the proposed data augmentation method, the performances of detection methods are improved consistently. Moreover, the experimental results show that the accuracy can be further improved by combining the proposed augmentation method with traditional data augmentation methods such as random scaling and cropping (see the last row).

\fref{fig:result} shows the qualitative results. We can observe that the proposed method localizes all three components while other methods often fail to localize small components. Localizing all components is important to be able to estimate both location and rotation. A failure case is shown in~\fref{fig:failure}. Images with significant motion blur or very dark scenes were found to be challenging.

An ablation study on the environment guide vector is shown in~\tref{tab:ablation}. For all the experiments except this table, the proposed data augmentation is performed by utilizing the vector that is randomly sampled from the uniform distribution $[-1, 1]$ as mentioned in~\sref{sec:detection}. For this ablation study, we consider the case that while ground truth annotations of charging inlets are not given for test images, we have access to those images. Then, while we cannot, of course, train detection methods using them, we can translate training images towards the distribution of them for data augmentation. The first row shows the result of using random sampling from uniform distribution. The other row is the result of utilizing the distribution of environments of validation and test images along with additive noises from uniform distribution.

We also investigate EnT-GAN-based Mosaic data augmentation in~\tref{tab:result_mosaic}. Given an original image in the training set, parts of the image are replaced by translated images as shown in~\fref{fig:result_mosaic}. Specifically, the number of selected regions for each image is between zero and four. The width/height of the region is 20$\sim$80\% of that of the image. The Mosaic images are utilized for training detection networks.

In this work, detection network and data augmentation network are trained separately. We believe that the performance can be further improved when they are trained concurrently.

% we randomly select four regions where the probability of replacing the region by the corresponding part of a translated image is $0.5$. The width/height of the region is the product of a value sampled from the uniform distribution $[0.2,0.8]$ times the width/height of the image, respectively. } 

\section{CONCLUSIONS}

Towards autonomous EV charging robots, we first introduce a novel dataset and present an experimental analysis of the existing methods on the dataset. Then, to improve the robustness of localization, we investigate a data augmentation method with a focus on controllable and explainable image-to-image translation. To achieve this, we propose to utilize an intuitive environment guide vector in the proposed image-to-image translation network. We demonstrate that the proposed method is able to successfully synthesize various and expected images given an image and environment vectors. The experimental results show that by utilizing the proposed data augmentation method to detection networks, their performances are improved on EV charging inlet localization.


%%%%%%%%%%%%%%%%%%%%%%%%%%%%%%%%%%%%%%%%%%%%%%%%%%%%%%%%%%%%%%%%%%%%%%%%%%%%%%%%

\bibliographystyle{IEEEtran}
\bibliography{mybibfile}

\begin{thebibliography}{10}
\providecommand{\url}[1]{#1}
\csname url@rmstyle\endcsname
\providecommand{\newblock}{\relax}
\providecommand{\bibinfo}[2]{#2}
\providecommand\BIBentrySTDinterwordspacing{\spaceskip=0pt\relax}
\providecommand\BIBentryALTinterwordstretchfactor{4}
\providecommand\BIBentryALTinterwordspacing{\spaceskip=\fontdimen2\font plus
\BIBentryALTinterwordstretchfactor\fontdimen3\font minus
  \fontdimen4\font\relax}
\providecommand\BIBforeignlanguage[2]{{%
\expandafter\ifx\csname l@#1\endcsname\relax
\typeout{** WARNING: IEEEtran.bst: No hyphenation pattern has been}%
\typeout{** loaded for the language `#1'. Using the pattern for}%
\typeout{** the default language instead.}%
\else
\language=\csname l@#1\endcsname
\fi
#2}}

\bibitem{Behl2019}
M.~Behl, J.~DuBro, T.~Flynt, I.~Hameed, G.~Lang, and F.~Park, ``Autonomous
  electric vehicle charging system,'' in \emph{2019 Systems and Information
  Engineering Design Symposium (SIEDS)}, 2019.

\bibitem{evChargingMiseikis2017}
J.~Miseikis, M.~R{\"u}ther, B.~Walzel, M.~Hirz, and H.~Brunner, ``3d vision
  guided robotic charging station for electric and plug-in hybrid vehicles,''
  in \emph{OAGM/AAPR ARW 2017: Joint Workshop on Vision, Automation, and
  Robotics}, 2017.

\bibitem{evChargingLong2019}
Y.~Long, C.~Wei, C.~Cao, X.~Hu, B.~Zhu, and F.~Long, ``Design of high-power
  fully automatic charging device,'' in \emph{2019 IEEE Sustainable Power and
  Energy Conference (iSPEC)}, 2019.

\bibitem{evChargingLou2020}
Y.~Lou and S.~Di, ``Design of a cable-driven auto-charging robot for electric
  vehicles,'' \emph{IEEE Access}, vol.~8, 2020.

\bibitem{evChargingSun2018}
C.~Sun, M.~Pan, Y.~Wang, J.~Liu, H.~Huang, and L.~Sun, ``Method for electric
  vehicle charging port recognition in complicated environment based on cnn,''
  in \emph{2018 15th International Conference on Control, Automation, Robotics
  and Vision (ICARCV)}, 2018.

\bibitem{houghTransform1972}
R.~O. Duda and P.~E. Hart, ``Use of the hough transformation to detect lines
  and curves in pictures,'' \emph{Commun. ACM}, vol.~15, no.~1, Jan. 1972.

\bibitem{augganECCV2018}
S.-W. Huang, C.-T. Lin, S.-P. Chen, Y.-Y. Wu, P.-H. Hsu, and S.-H. Lai,
  ``Auggan: Cross domain adaptation with gan-based data augmentation,'' in
  \emph{Computer Vision -- ECCV 2018}, 2018.

\bibitem{augganTITS2021}
C.-T. Lin, S.-W. Huang, Y.-Y. Wu, and S.-H. Lai, ``Gan-based day-to-night image
  style transfer for nighttime vehicle detection,'' \emph{IEEE Transactions on
  Intelligent Transportation Systems}, vol.~22, no.~2, 2021.

\bibitem{augganAAAI2020}
C.-T. Lin, Y.-Y. Wu, P.-H. Hsu, and S.-H. Lai, ``Multimodal
  structure-consistent image-to-image translation,'' \emph{Proceedings of the
  AAAI Conference on Artificial Intelligence}, vol.~34, no.~07, Apr. 2020.

\bibitem{nightDALeeAccess2020}
H.~Lee, M.~Ra, and W.-Y. Kim, ``Nighttime data augmentation using gan for
  improving blind-spot detection,'' \emph{IEEE Access}, vol.~8, 2020.

\bibitem{forkGANECCV2020}
Z.~Zheng, Y.~Wu, X.~Han, and J.~Shi, ``Forkgan: Seeing into the rainy night,''
  in \emph{Computer Vision -- ECCV 2020}, 2020.

\bibitem{cycleGAN2017}
J.-Y. Zhu, T.~Park, P.~Isola, and A.~A. Efros, ``Unpaired image-to-image
  translation using cycle-consistent adversarial networks,'' in \emph{2017 IEEE
  International Conference on Computer Vision (ICCV)}, 2017.

\bibitem{dayToNightAnokhinCVPR2020}
I.~Anokhin, P.~Solovev, D.~Korzhenkov, A.~Kharlamov, T.~Khakhulin,
  A.~Silvestrov, S.~Nikolenko, V.~Lempitsky, and G.~Sterkin, ``High-resolution
  daytime translation without domain labels,'' in \emph{2020 IEEE/CVF
  Conference on Computer Vision and Pattern Recognition (CVPR)}, 2020.

\bibitem{nightToDayAshaICRA2019}
A.~Anoosheh, T.~Sattler, R.~Timofte, M.~Pollefeys, and L.~V. Gool,
  ``Night-to-day image translation for retrieval-based localization,'' in
  \emph{2019 International Conference on Robotics and Automation (ICRA)}, 2019.

\bibitem{RCNN2014}
R.~Girshick, J.~Donahue, T.~Darrell, and J.~Malik, ``Rich feature hierarchies
  for accurate object detection and semantic segmentation,'' in \emph{2014 IEEE
  Conference on Computer Vision and Pattern Recognition}, 2014.

\bibitem{fastRCNN2015}
R.~Girshick, ``Fast r-cnn,'' in \emph{2015 IEEE International Conference on
  Computer Vision (ICCV)}, 2015.

\bibitem{fasterRCNN2015}
S.~Ren, K.~He, R.~Girshick, and J.~Sun, ``Faster r-cnn: Towards real-time
  object detection with region proposal networks,'' in \emph{Advances in Neural
  Information Processing Systems}, vol.~28, 2015.

\bibitem{fasterRCNN2017}
------, ``Faster r-cnn: Towards real-time object detection with region proposal
  networks,'' \emph{IEEE Transactions on Pattern Analysis and Machine
  Intelligence}, vol.~39, no.~6, 2017.

\bibitem{yolo2016}
J.~Redmon, S.~Divvala, R.~Girshick, and A.~Farhadi, ``You only look once:
  Unified, real-time object detection,'' in \emph{2016 IEEE Conference on
  Computer Vision and Pattern Recognition (CVPR)}, 2016.

\bibitem{ssd2016}
W.~Liu, D.~Anguelov, D.~Erhan, C.~Szegedy, S.~Reed, C.-Y. Fu, and A.~C. Berg,
  ``Ssd: Single shot multibox detector,'' in \emph{Computer Vision -- ECCV
  2016}, 2016.

\bibitem{retinaNet2017}
T.-Y. Lin, P.~Goyal, R.~Girshick, K.~He, and P.~Dollár, ``Focal loss for dense
  object detection,'' in \emph{2017 IEEE International Conference on Computer
  Vision (ICCV)}, 2017.

\bibitem{efficientDet2020}
M.~Tan, R.~Pang, and Q.~V. Le, ``Efficientdet: Scalable and efficient object
  detection,'' in \emph{2020 IEEE/CVF Conference on Computer Vision and Pattern
  Recognition (CVPR)}, 2020.

\bibitem{efficientNet2019}
M.~Tan and Q.~Le, ``{E}fficient{N}et: Rethinking model scaling for
  convolutional neural networks,'' in \emph{Proceedings of the 36th
  International Conference on Machine Learning}, vol.~97, 2019.

\bibitem{detectoRS2020}
S.~Qiao, L.-C. Chen, and A.~Yuille, ``Detectors: Detecting objects with
  recursive feature pyramid and switchable atrous convolution,'' in
  \emph{Proceedings of the IEEE/CVF Conference on Computer Vision and Pattern
  Recognition (CVPR)}, June 2021.

\bibitem{HTC2019}
K.~Chen, J.~Pang, J.~Wang, Y.~Xiong, X.~Li, S.~Sun, W.~Feng, Z.~Liu, J.~Shi,
  W.~Ouyang, C.~C. Loy, and D.~Lin, ``Hybrid task cascade for instance
  segmentation,'' in \emph{2019 IEEE/CVF Conference on Computer Vision and
  Pattern Recognition (CVPR)}, 2019.

\bibitem{transformer2017}
A.~Vaswani, N.~Shazeer, N.~Parmar, J.~Uszkoreit, L.~Jones, A.~N. Gomez, L.~u.
  Kaiser, and I.~Polosukhin, ``Attention is all you need,'' in \emph{Advances
  in Neural Information Processing Systems}, vol.~30, 2017.

\bibitem{DETR2020}
N.~Carion, F.~Massa, G.~Synnaeve, N.~Usunier, A.~Kirillov, and S.~Zagoruyko,
  ``End-to-end object detection with transformers,'' in \emph{Computer Vision
  -- ECCV 2020}, 2020.

\bibitem{deformableDETR2021}
X.~Zhu, W.~Su, L.~Lu, B.~Li, X.~Wang, and J.~Dai, ``Deformable detr: Deformable
  transformers for end-to-end object detection,'' \emph{arXiv}, 2021.

\bibitem{DETRACT2021}
M.~Zheng, P.~Gao, X.~Wang, H.~Li, and H.~Dong, ``End-to-end object detection
  with adaptive clustering transformer,'' in \emph{32nd British Machine Vision
  Conference (BMVC)}, 2021.

\bibitem{imgProcessingToNight2002}
W.~B. Thompson, P.~Shirley, and J.~A. Ferwerda, ``A spatial post-processing
  algorithm for images of night scenes,'' \emph{Journal of Graphics Tools},
  vol.~7, no.~1, 2002.

\bibitem{rgb2gray}
``Bt.601-7: Studio encoding parameters of digital television for standard 4:3
  and wide screen 16:9 aspect ratios,'' \emph{ITU-R}, 2011.

\bibitem{adain}
X.~Huang and S.~Belongie, ``Arbitrary style transfer in real-time with adaptive
  instance normalization,'' in \emph{2017 IEEE International Conference on
  Computer Vision (ICCV)}, 2017.

\bibitem{lsgan}
X.~Mao, Q.~Li, H.~Xie, R.~Y. Lau, Z.~Wang, and S.~P. Smolley, ``Least squares
  generative adversarial networks,'' in \emph{2017 IEEE International
  Conference on Computer Vision (ICCV)}, 2017.

\bibitem{lin2014microsoft}
T.-Y. Lin, M.~Maire, S.~Belongie, J.~Hays, P.~Perona, D.~Ramanan,
  P.~Doll{\'a}r, and C.~L. Zitnick, ``Microsoft coco: Common objects in
  context,'' in \emph{Computer Vision -- ECCV 2014}, 2014.

\end{thebibliography}

\end{document}